\documentclass{article}

\usepackage[preprint]{neurips_2023}

\usepackage[utf8]{inputenc} 
\usepackage[T1]{fontenc}    
\usepackage{hyperref}       
\usepackage{url}            
\usepackage{booktabs}       
\usepackage{amsfonts}       
\usepackage{nicefrac}       
\usepackage{microtype}      
\usepackage{xcolor}         
\usepackage{bbm}

\usepackage[inkscapeformat=png]{svg}

\usepackage{graphicx} 
\usepackage{algorithm}
\usepackage{algpseudocode}
\usepackage{dsfont}
\usepackage{amsmath}
\usepackage{subfig}
\usepackage{wrapfig}
\usepackage{comment}
\usepackage{amssymb,amsfonts,bm}

\newcommand{\CD}{\mathcal{D}}

\newcommand{\ov}{\bm{1}}

\newcommand{\tr}{^{\intercal}}

\definecolor{b_color}{HTML}{F05039}
\definecolor{u_color}{HTML}{1F449C}

\title{Fixing confirmation bias in feature attribution methods via semantic match}

\author{%
Giovanni Cin\`{a}$^{1, 2}$ \quad Daniel Fernández-Llaneza$^{1}$ \quad \textbf{Ludovico Deponte}$^{2}$\quad Nishant Mishra$^1$ \\ \textbf{Tabea E. R\"ober}$^3$ \quad \textbf{Sandro Pezzelle}$^2$ \quad \textbf{Iacer Calixto}$^1$ \quad
\textbf{Rob Goedhart}$^3$ \quad \textbf{\c{S}. \.{I}lker Birbil}$^3$ \quad 
\\
$^1$Department of Medical Informatics, Amsterdam University Medical Center, \\ Amsterdam Public Health, Methodology \& Mental Health, Amsterdam, The Netherlands \\
$^2$Institute for Logic, Language and Computation, University of Amsterdam, \\Amsterdam, The Netherlands \\ $^3$Department of Business Analytics, Amsterdam Business School, University of Amsterdam \\Amsterdam, The Netherlands \\
\texttt{\{g.cina,d.fernandezllaneza,n.mishra,i.coimbra\}@amsterdamumc.nl}\\
\texttt{\{t.e.rober,s.pezzelle,r.goedhart2,s.i.birbil\}@uva.nl}\\ \texttt{ludovico.deponte@gmail.com}
}

\begin{document}

\maketitle

\begin{abstract}
Feature attribution techniques have become a staple method to disentangle the complex behavior of black box models. Despite the success of these methods, they suffer from a serious flaw: visualizing feature contributions is not enough for humans to reliably conclude something about a model's internal representations, and confirmation bias can trick users into false beliefs about model behavior. We propose a structured approach to test whether our hypotheses on the model are confirmed by feature attributions, in order to prevent confirmation bias. We showcase the procedure in a suite of experiments spanning tabular, image and textual
  data, and demonstrate how the assessment of semantic match can give insight into both desirable (\textit{e.g.}, focusing on an object relevant for prediction) and undesirable (\textit{e.g.}, focusing on a spurious correlation) model behaviors. The experiments range from synthetic data and tasks to real-world data. 
  We couple our results with an analysis on some proposed metrics to measure semantic match. 
\end{abstract}

\section{Introduction}
\label{sec:intro}

The success of machine learning techniques in solving a variety of tasks, along with  a parallel surge in model complexity, has rekindled interest in the interface between humans and machines. The field of Explainable AI (XAI) is concerned with unpacking the complex behavior of machine learning models in a way that is digestible by humans \citep[]{doshi-velez2017, linardatos2020, Gilpin.2018, biran2017, doran2017}. 

Among several proposed solutions, one approach has risen to prominence in the last half decade: feature attribution, also known as feature importance. In a nutshell, feature attribution methods explain model behavior by indicating the extent to which different parts of the input contribute to the model's output. These methods are currently employed in a plethora of scenarios, including virtually all data modalities, and deployed in production in low- as well as high-risk environments \citep{bhatt2020, thoral2021explainable}. 

Yet, such techniques are not free from criticism. Besides raising doubts about consistency between explanations and model behavior, scholars have argued that feature attribution techniques expose the users of machine learning applications to confirmation bias, namely the reasoning pitfall that leads us to believe an explanation just because it aligns with our expectations \citep{lipton2018, Ghassemi.2021}. For instance, a clinician using AI to diagnose metabolic disorders from images---after inspecting some explanations highlighting build-ups of fat in the liver---might be prone to believe that the model has learned to detect fatty liver. As fatty liver is a known metabolic condition, the clinician recognizes it and possibly assumes that the machine does, too. This may influence the level of trust the clinician has in the model, affecting the way care is delivered. But how can we be sure the model has learned this clinical insight? 

More generally, feature attributions are sub-symbolic in nature: they represent feature importance as vectors in $\mathbb{R}^n$ or matrices in $\mathbb{R}^n\times \mathbb{R}^m$, for some $n,m$. We currently have no systematic way to ascertain whether such explanations capture a concept we are interested in. Some authors advocate checking explanations against human intuition \citep{neely2022song}, but this exercise must be structured in a way that allows us to \textit{measure} alignment between human concepts and explanations, lest we fall back into the problem of confirmation bias. 

In this article, we build on the framework of semantic match proposed by \citet{cina2023semanticmatch} and make use of a machinery similar to that of \citet{zhou2022exsum} to formalize a procedure that allows us (1) to formulate a hypothesis of the form ``the model behaves in this way,'' (2) to obtain a score representing the extent to which the model's explanations confirm or reject this hypothesis, and (3) to obtain a measure of how much explanations discriminate situations when the model adheres (or not) to the hypothesis.
Such procedure is general and in principle can be applied to any model and to any local feature attribution method. Here, we focus on SHapley Additive exPlanations (SHAP) \citep{lundberg2017} due to their widespread use in practice. This methodology is paired with a discussion on what metrics are appropriate to measure semantic match. 
We display the procedure with experiments on tabular, image, and text data. To validate the method against a ground truth, we test the semantic match of hypotheses concerning properties of models that are known in the literature.
Furthermore, we investigate different kinds of hypotheses about previously unknown model behavior, showing that the procedure can give insight into  both desirable --\textit{i.e.}, what we hope the model is doing well-- as well as undesirable behaviors. All experiments use publicly available data and are fully reproducible\footnote{\url{https://github.com/Giovannicina/semantic_match}}.


\section{Related Work}
\label{sec:relatedwork}

\textbf{Feature attribution methods in XAI.} The majority of attribution-based methods provide local explanations, \textit{i.e.}, they aim to explain the prediction for an individual instance, hence we focus our work on local attribution-based methods. The latter can be divided into gradient-based and perturbation-based methods.
The former category includes, for instance, DeConvNet \citep{zeiler2014visualizing}, guided backpropagation (GBP) \citep{springenberg2014striving}, Grad-CAM \citep{selvaraju2017grad}, and integrated gradients \citep{sundararajan2017axiomatic}. 
Methods that fall in the second category include occlusion sensitivity maps \citep{zeiler2014visualizing}, LIME \citep{Ribeiro.2016} and SHAP \citep{lundberg2017}. We refer to \citet{abhishek2022attribution} and \citet{jimaging6060052} for a more detailed account of different methods.

\textbf{Confirmation bias in XAI.} 
Confirmation bias is a well-known concept from psychology, first described by \citet{wason1960_confirmationbias}, and followed by plenty of empirical investigations  \citep[\textit{e.g.}][]{lord1979biased, evans1989bias, nickerson1998_confirmationbias}. The American Psychological Association defines confirmation bias as ``the tendency to gather evidence that confirms preexisting expectations, typically by emphasizing or pursuing supporting evidence while dismissing or failing to seek contradictory evidence'' \citep{American_Psychological_Association}. Even though the problem of this type of cognitive bias has been acknowledged in the (X)AI literature \citep[\textit{e.g.,}][]{Ghassemi.2021, cina2023semanticmatch, rudin_stop2019}, the empirical research on confirmation bias in XAI is scarce. \citet{Wang2019_usercentricXAI} propose a conceptual framework for building human-centered and decision-theory-driven XAI, in which they consider human decision making and the role of confirmation bias in relation to XAI. \citet{wan2022bias} conducted a field experiment in which subjects were tasked with performing risk assessments aided by a predictive model, while
\citet{Bauer2023_XAI} conducted two studies in the real estate industry investigating how humans shift their mental models. 
Both results find that confirmation bias is present in human-XAI interaction.



The risk of falling prey to confirmation bias is especially present if we use feature attribution methods on high-level features \citep{cina2023semanticmatch}. Particularly in the field of computer vision and deep learning, \textit{high-level} features refer to patterns in groups of features, while \textit{low-level} features are the entries of the input vector \citep[\textit{e.g.,}][]{zeiler2014visualizing, Lee_2016_CVPR, Deng2014}.
\citet{cina2023semanticmatch} argue that the meaning of feature attributions for low-level features is intuitive, if the low-level features have a predefined semantic translation, as in most tabular data. In image data, however, individual pixels do not carry any semantic meaning. Hence, the use of feature attribution methods is not sensible, unless we know whether the high-level features match our semantic representation, \textit{i.e.}, if we have \emph{semantic match} \citep{cina2023semanticmatch, pmlr-v80-kim18d}.

\textbf{Relevant approaches.} 
In Natural Language Processing (NLP) a similar approach is \textit{probing classifiers}, a way of understanding whether a language model's internal representation is encoding some linguistic property \citep{belinkov2022probing, hupkes2018visualisation}. Despite the shared intention to unpack sub-symbolic representations, model embeddings are not explanations and probing does not appeal to intuitions in the same way as feature attribution methods do.  
In image classification, a typical approach to explain the classification is using prototypes, namely explanations providing prototypical images from the training data to explain some classification \citep[\textit{e.g.,}][]{arik2020prototypes, biehl2016prototypes, nauta2021}. Another approach for interpretable image classification is concept-based models or concept bottleneck models (CBMs) \citep[\textit{e.g.,}][]{pmlr-v80-kim18d, Barbiero_Ciravegna_Giannini_Lió_Gori_Melacci_2022, Yuksekgonul2022CBM, pmlr-v202-ghosh23c}. The core idea of such approaches is to map inputs onto some user-defined concepts, which are then used to predict the outcome class. In both CBMs and prototype explanations, the intention is to ground explanations by latching them to concepts or prototypes for which semantic match is given. Finally, \citet{zhou2022exsum} formalize and measure the validity of hypotheses such as `all positive adjectives are assigned positive contribution.' This formalism provides the closest attempt to what we aim to achieve, with the crucial difference that hypotheses are still formulated at the level of low-level features, specifically tokens (\textit{e.g.}, a positive adjective) with a clear semantic interpretation, while our work offers a more general approach able to handle hypotheses on high-level features, which are the main source of confirmation bias.

\textbf{Contributions.} We propose an approach to formulate  hypotheses on model behavior and test directly whether semantic match is present. Our metrics for semantic match support a quantitative analysis of the explanations, allowing us to sidestep our intuition and avoid confirmation bias. 

\section{Methodology}
\label{sec:methods}

In this section we describe the main methodology in full generality, elaborate on the metrics to assess semantic match and outline the setup of the experiments.

\subsection{Defining a hypothesis}

Consider a dataset $\CD = \{(\bm{x}_i, y_i) : i=1, \dots, n\}$ where $\bm{x}_i$ is the input vector and $y_i$ is the label of sample $i$. We assume that a machine learning model $f$ is trained on this dataset, and then, a local attribution method $M$ is specified. The term $M(f, \bm{x}_i, y_i) = \bm{e}_i$ defines the explanation vector $\bm{e}_i$ obtained by $M$ for data point $i$ classified with model $f$. 

We want to formulate hypotheses as logical statements that can be quantitatively evaluated. 
For example, suppose we have a model classifying road images into safe or dangerous situations. Let us consider how to implement the hypothesis $\theta$ = `when there is a pedestrian in the image and a zebra crossing, the model attributes considerable importance to the zebra crossing'.
In this case the zebra crossing is a high-level feature that (i) may not correspond to a single low-level input feature and (ii)  may correspond to different sets of pixels in different images. Our hypotheses concern how the model behaves on tuples $\bm{u}_i = (f, \bm{x}_i, y_i)$ consisting of a model, a data point, and its label. For a model $f$, call the set of all such tuples $\mathcal{U}$. We partially follow \citet{zhou2022exsum} in defining a \textit{hypothesis} $\theta$ in the logical form \textit{if $A_\theta$ then $B_\theta$}, where $A_\theta$ describes the \emph{applicability} (the tuples that the hypothesis applies to) and $B_\theta$ describes the \emph{behavior} (expected explanations). We drop the subscript whenever clear. We write $ \bm{u}_i\models_A A_\theta$ and $\bm{e}_i \models_B B_\theta$ to indicate the compliance with the applicability and behavior conditions, respectively. A tuple $\bm{u}_i$ \textit{satisfies} a hypothesis $\theta$ if both $ \bm{u}_i\models_A A_\theta$ and $M(\bm{u}_i) = \bm{e}_i \models_B B_\theta$ hold; we denote this with the shorthand $\bm{u}_i\models \theta$.  For our example, say we have bounding boxes for both pedestrians and zebra crossings in all images, if they appear. We could stipulate that $ \bm{u}_i\models A_\theta$, if the image $x_i$ has non-empty bounding boxes for both a pedestrian and a zebra crossing. Condition $B$ could be implemented as $\bm{e}_i \models B_\theta$ iff the bounding box for the zebra crossing contains at least 80\% of the contribution to the outcome according to $\bm{e}_i$ (normalized for the whole image).

\subsection{Testing hypothesis}

Now that we are able to specify a hypothesis, the natural question to ask is: is this hypothesis valid? \citet{zhou2022exsum} define \textit{validity} to be the expected proportion of tuples satisfying $A$ whose behavior complies to $B$:\footnote{Additionally, \citet{zhou2022exsum} define coverage as $ P(\bm{u}_i\models A_\theta)$, namely the proportion of tuples triggering the hypothesis and sharpness  as  $P(\bm{u}_i\models A_\theta|M(\bm{u}_i)\models B_\theta)$.}
$$val(\theta)= P(M(\bm{u}_i)\models B_\theta | \bm{u}_i\models A_\theta)$$

\begin{figure}[h]
    \centering
    \includegraphics[width=0.45\columnwidth]{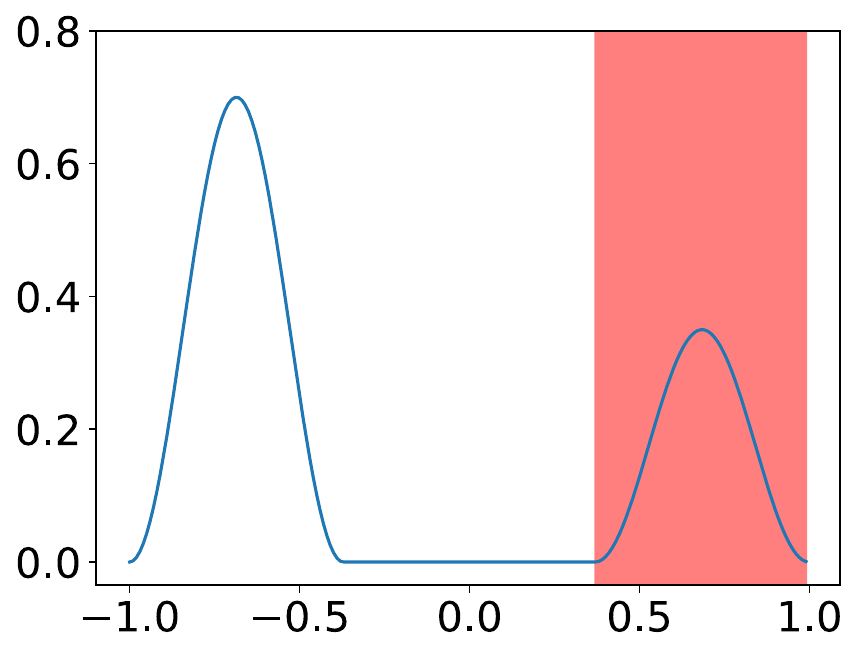}
    \includegraphics[width=0.45\columnwidth]{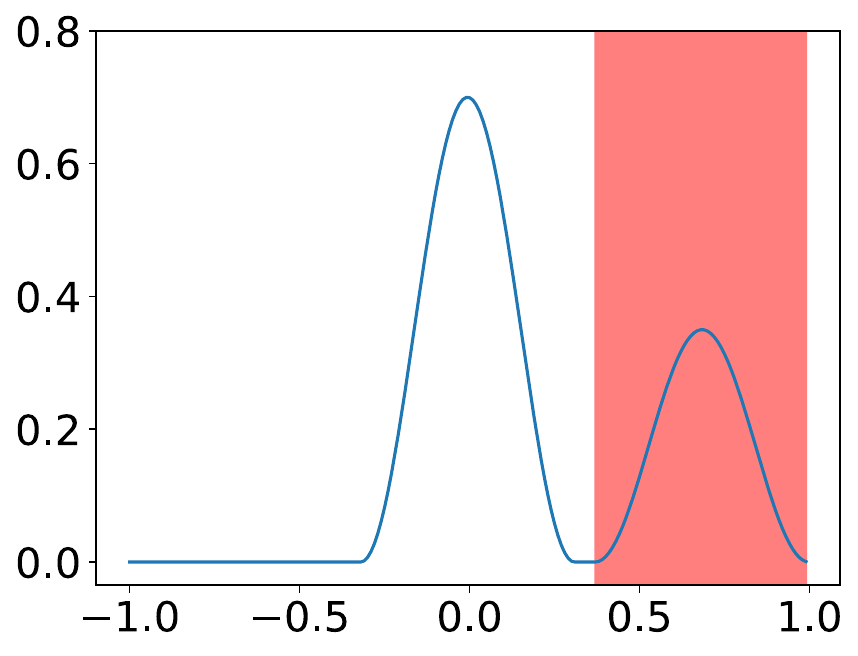} 
    \caption{Two histograms representing the distribution of explanations for a specific feature, for all points satisfying condition $A$. When assessing the validity of a hypothesis with condition $\bm{e} \models B$ iff $\bm{e}\in [0.3, 1]$, the hypothesis will have the same validity in both cases. However, the distribution on the left is less coherent with $\theta$ compared to the distribution on the right.} 
\label{fig:distributions}
\end{figure}

This definition is intuitive but suffers from the formulation of $B$ in terms of a sharp threshold (like the 80\% of the previous example). Suppose we have two different sets of explanations, each explanation being a number in $[-1, 1]$, and that our hypothesis concerns a single feature. We can visualize the two sets of explanations for this feature with two histograms as in Figure \ref{fig:distributions}. A hypothesis concerning a behavior in the range depicted will have the same validity in both scenarios. Yet, the scenario on the right is more coherent with the hypothesis, because the mass of the explanations is closer to the hypothesized cutoff point.

This points to two further questions: 1) For points satisfying precondition $A$, are explanations \textit{coherent} in complying with $\theta$ or are there starkly different behaviours? 2) Are the explanations \textit{discriminative} for the hypothesis $\theta$ meaning that they allow us to distinguish when the model complies with $\theta$ and when not?
We address these in turn by defining specific metrics.

\looseness=-1
\textbf{Testing for coherence. }
Coherence between explanations and hypothesis means that that all the data points giving rise to similar explanations are also complying with our hypothesis $\theta$, and vice versa \citep{cina2023semanticmatch}. 
We assess similarity by means of a distance between explanations and write the distance between the explanations of data points $i$ and $c$ as $d(\bm{e}_i, \bm{e}_c)$. 
In case of explanations as vectors or matrices, there are standard notions of distance to employ; see Section \ref{sec:results} for  examples. A hypothesis has a good semantic match when all explanations are close to the explanations of tuples satisfying $\theta$. To quantify this, we employ the median distance of explanations from a reference point with index $c$:
\begin{equation}
\label{mediandistance}
    MD(\theta, \bm{e}_c) = \text{median}\{d(M(\bm{u}_i), \bm{e}_c)~|~ \bm{u}_i\in \mathcal{U}\}. 
\end{equation}
The dependence on the reference point can be sidestepped by re-sampling it from the set of points satisfying the hypothesis. In Figure \ref{fig:distributions}, median distance sets apart the two cases considered: the distribution on the left would have a higher median distance, signaling a worse match.

\textbf{Testing for discrimination. }
A hypothesis might hold for specific subgroups of tuples, and not for others. In this scenario, can explanations help us distinguish tuples that satisfy $\theta$ from those that do not? To answer this question we can use the distances defined above to \textit{predict} whether a tuple satisfies the hypotheses.
We construct a new dataset by relabeling the tuples using the hypothesis $\theta$. That is, for all $\bm{u}_i\models_A A_\theta$ we obtain new labels with $\theta$ by defining 
$\bar{y}_i = \mathds{1}\big( \bm{u}_i  \models \theta \big)$.
Here, $\mathds{1}$ stands for the indicator function. 
For all such tuples, we can then use distances $d(M(\bm{u}_i), \bm{e}_c)$ to predict such labels.
When considering the distances as a ranking, we need to flip the sign since in standard classification problems larger values denote the positive class. With this approach, we can resort to well-known metrics to measure the discrimination of rankings, such as the area under the ROC curve (shortened with AUC). 


\subsection{Experimental setup}
\label{sec:setup}

\textbf{Tabular data. }
We first design a controlled experiment on synthetic data so that we can have clear expectations on whether semantic match should work or not.
We generated a tabular dataset consisting of two normally distributed continuous features, $x_1$ and $x_2$, and one binary feature $x_3$. We proceeded to define a binary outcome by passing the function $x_1x_3 - (1-x_3)x_1 + x_2$ through a sigmoid and a 0.5 threshold. In this way, we incorporate a feature interaction between features $x_1$ and $x_3$ into the outcome: this interaction will be the high-level feature of interest.
We then trained a random forest on the dataset in order to predict the outcome, and generated explanations using SHAP. 
We formulated the following hypothesis $\theta$ to express that the model has learned part of the interaction:  ``when $x_1$ is negative and $x_3 = 0$, the model assigns positive contribution to $x_1$''. Formally for sample $i$, we have
\begin{align*}
    \bm{u}_i \models_A A &\iff (x_{i1} < 0) \land (x_{i3} = 0),\\
    \bm{e}_i \models_B B &\iff (e_{i1} > 0).
\end{align*}
Finally, we define a notion of distance between explanations. We opted for Euclidean distance between vectors of SHAP values: $d(\bm{e}_i, \bm{e}_c) = \|\bm{e}_i-\bm{e}_c\|$. With these ingredients we are then able to test for semantic match.



\textbf{MALeViC dataset. } 
MALeViC \citep{pezzelle-fernandez-2019-red} is 
a dataset of synthetically-generated images depicting four to nine colored geometric shapes with varying areas. The shapes are generated at random locations. Each shape has a corresponding binary label---\emph{big} or \emph{small}---which indicates whether it counts as big or small given the surrounding objects. These are based on an underlying 
threshold function considering the area occupied by the objects; see Appendix \ref{appendix:image_data_exp} for details. 
To solve this task, a model needs to construct high-level features capturing the size of the target object in relationship with the other shapes.

We focus on the partition of the MALeViC dataset where all the objects in an image are
either squares or rectangles.
This choice has a practical motivation, namely to have a direct mapping between objects and their bounding boxes. Furthermore, we select images that contain one single red object. The resulting dataset is balanced in terms of objects' sizes. We split our dataset into training, validation, and test sets (80:10:10). To augment our training data, we flip each image horizontally and vertically. Thus, we end up with 4,800 images in the training set and 200 images in the test set.
We trained a convolutional neural network (CNN) to predict whether red objects are \textit{big} or \textit{small}. Note that, beside the image, the model has no additional information on which objects needs to be classified. Further details on the implementation are specified in Appendix \ref{appendix:image_data_exp}.

\begin{figure}[ht!]
    \centering
    \includegraphics[width=1.0\columnwidth]{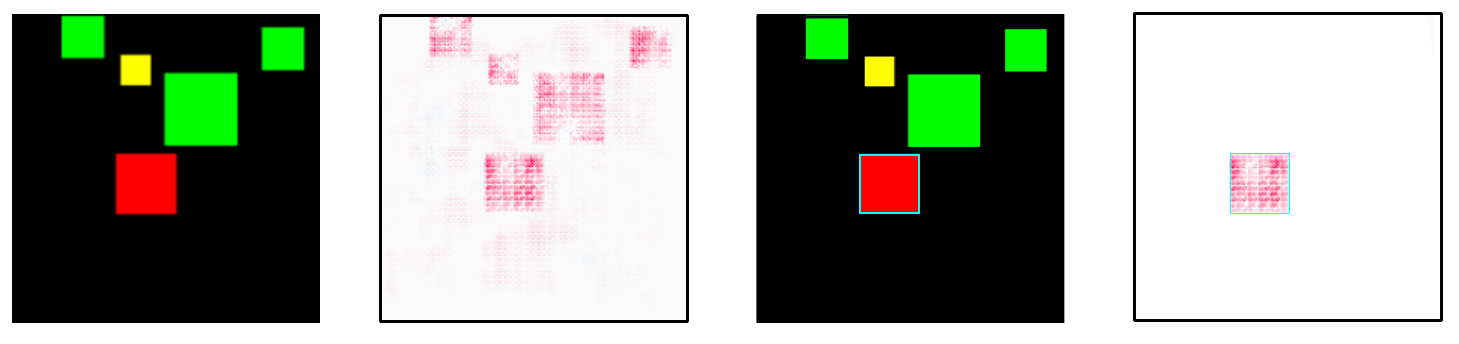}
     \vspace*{-5mm}
    \caption{An example image from the MALeViC dataset (left), alongside the SHAP values generated by the model (center-left), the \textcolor{cyan}{cyan} bounding box around the object of interest (center-right) and the SHAP values after masking is applied (right).}
\label{fig:example_image}
\end{figure}

All the hypotheses considered concern the contribution placed on specific objects in the image. It is however difficult to compare heatmaps directly because the shapes spawn at random locations in the images. We construct image-specific masks by segmenting the whole image with Segment Anything Model (SAM) \citep{kirillov2023segment} and by matching their coordinates to the relevant bounding boxes in the MALeVIC metadata. Thus, we calculate the amount of SHAP values placed on the relevant shape using the bounding box as a mask on the SHAP heatmap (see Figure \ref{fig:example_image}). We add up the absolute value of the contribution of the pixels within the bounding box and normalize this quantity by the total contribution in the whole image. This gives us the proportion of the contribution placed on the target object in the reference image $\bm{x}_c$ according to SHAP. 
For instance, we have the following hypothesis: ``if the prediction is correct, then we expect at least 10\% of the contribution on the red target object.'' Formally,
\[
\begin{array}{ll}
    \bm{u}_i \models_A A &\iff (y_i = f(\bm{x}_i)),\\
    \bm{e}_i \models_B B &\iff \left(\frac{\ov\tr \bm{M}_i \bm{e}_i}{\ov\tr \bm{e}_i} \geq 0.1\right),
\end{array}
\]
where the matrix $\bm{M}_i$ is the masking operator (also known as bounding box) used to designate the target object in the sample image $i$ and $\ov\tr \bm{M}_i \bm{e}_i$ is matrix notation to denote the sum of SHAP contributions on a certain area.
We want a notion of distance that only considers the relevant parts of the images. Given two images, we calculate the proportion of contribution of the red shape in both and measure the distance as the absolute difference in these proportions:
\[
d(\bm{e}_i, \bm{e}_c) = \left| \frac{\ov\tr \bm{M}_i \bm{e}_i}{\ov\tr \bm{e}_i} - \frac{\ov\tr \bm{M}_c \bm{e}_c}{\ov\tr \bm{e}_c}\right|.
\]
Hence, if the contribution on the red target object is 40\% in our reference explanation and 10\% in another explanation, the distance between the two is 30\% or 0.3. 

\textbf{VOC2006 dataset. }
To test our setting in a real-world dataset, we utilized the PASCAL Visual Object Classes Challenge 2006 (VOC2006) dataset \citep{everingham2010pascal}. It contains 5,307 images, comprising 9,507 annotated objects. The objects can be truncated due to occlusion by another object or because the object extends outside the image margins, and can be placed in different views: frontal, rear, left, or right. The objects can be classified as vehicles (\textit{i.e.}, bicycle, bus, car, motorbike) or mammals (\textit{e.g.},  cow, dog, horse, person, sheep), and each of them comes equipped with a rectangular bounding box.
We used the sets provided by default for training (1,277 images, 25\%), validation (1,341 images, 25\%), and test (2,686, 50\%) and assigned a binary label to the image depending on whether there is a vehicle or not, which the CNN was trained to predict. The training images are augmented with horizontal flips yielding a total of 2,554 training images. The CNN architecture was the same as in the MALeViC experiments (see Figure \ref{fig:architecture} in Appendix \ref{appendix:image_data_exp}). 
The explanations were also generated using pixel-level SHAP values based on the bounding boxes provided (see Figure \ref{fig:shap_cars_bbox} in Appendix \ref{appendix:image_data_exp_voc}). All hypotheses considered concern the contribution placed on the target (\textit{i.e.}, vehicle or mammal) objects of each image and mimic the ones presented for the MALeViC dataset.

\textbf{SQuAD dataset. } 
We also include an experiment on a common NLP task, question answering, on the SQuAD dataset. \citet{ko2021look} showed that it is possible to build biased models which focus only on specific parts of the context, by training them on data subsets that only contain answers in the $n$th sentence of the context.
With this experiment, we want to employ semantic match to discriminate between a biased and an unbiased model.
We investigate bias in the first sentence, as the subdataset containing only answers in the first sentence is the largest. Using a similar pipeline as \citet{ko2021look}, we fine-tune two BERT-base uncased models \citep{devlin2019bert}; one on a biased dataset where the answer is always in the first sentence, and one on a dataset with approximately the same number of data points but with answers spread evenly among sentences. 
Details on implementation, as well as evidence on the fact that the first model is indeed biased, can be found in Appendix \ref{appendix:squad_exp}. 
In this task, the BERT model outputs two independent probability distributions over the tokens in the context: one for the starting position and one for the ending position of the answer. Thus, we have SHAP values for both positions.
We map each token in a sample to the corresponding sentence using spaCy sentencizer \citep{spacy2023}. Analogous to the image experiment, one can think of the (pre-image of the) mapping as drawing a bounding box around each sentence. 
We process the SHAP values to obtain the percentage of contribution of each sentence for the prediction (see Appendix \ref{appendix:squad_exp}).

All our hypotheses focus on the relation between model's SHAP contribution, answer position, and correctness of prediction. The goal is to capture the biased behavior, so we formulate the hypotheses in terms of contribution placed by the model on the first sentence. For instance, we have the hypothesis: ``if the prediction is correct and the answer is in the first sentence, then we expect at least 30\% of the contribution on the first sentence.'' Formally,
\[
\begin{array}{ll}
    \bm{u}_i \models_a A &\iff (y_i = f(x_i)) \land R(\bm{x}_i) = 1,\\
    \bm{e}_i \models_b B &\iff \left(\frac{\ov\tr \bm{M}_i \bm{e}_i}{\ov\tr \bm{e}_i} > 0.3\right),
\end{array}
\]
where the operator $R$ returns the index of the sentence bearing the answer, and the matrix $\bm{M}_i$ is the masking operator used to evaluate the contribution of the the first sentence for sample $i$. 
Again, as distance we use the absolute difference between the contribution percentage on the first sentence of the reference point and that of the other data points:
\[
d(\bm{e}_i, \bm{e}_c) = \left| \frac{\ov\tr \bm{M}_i \bm{e}_i}{\ov\tr \bm{e}_i} - \frac{\ov\tr \bm{M}_c \bm{e}_c}{\ov\tr \bm{e}_c}\right|.
\]
For example, if one explanation places $50\%$ of the average contribution on the first sentence and the other explanation  only $5\%$, the distance is $0.45$. 

\section{Results}
\label{sec:results}

\begin{figure}[h]
    \centering
    \includegraphics[width=0.7\columnwidth]{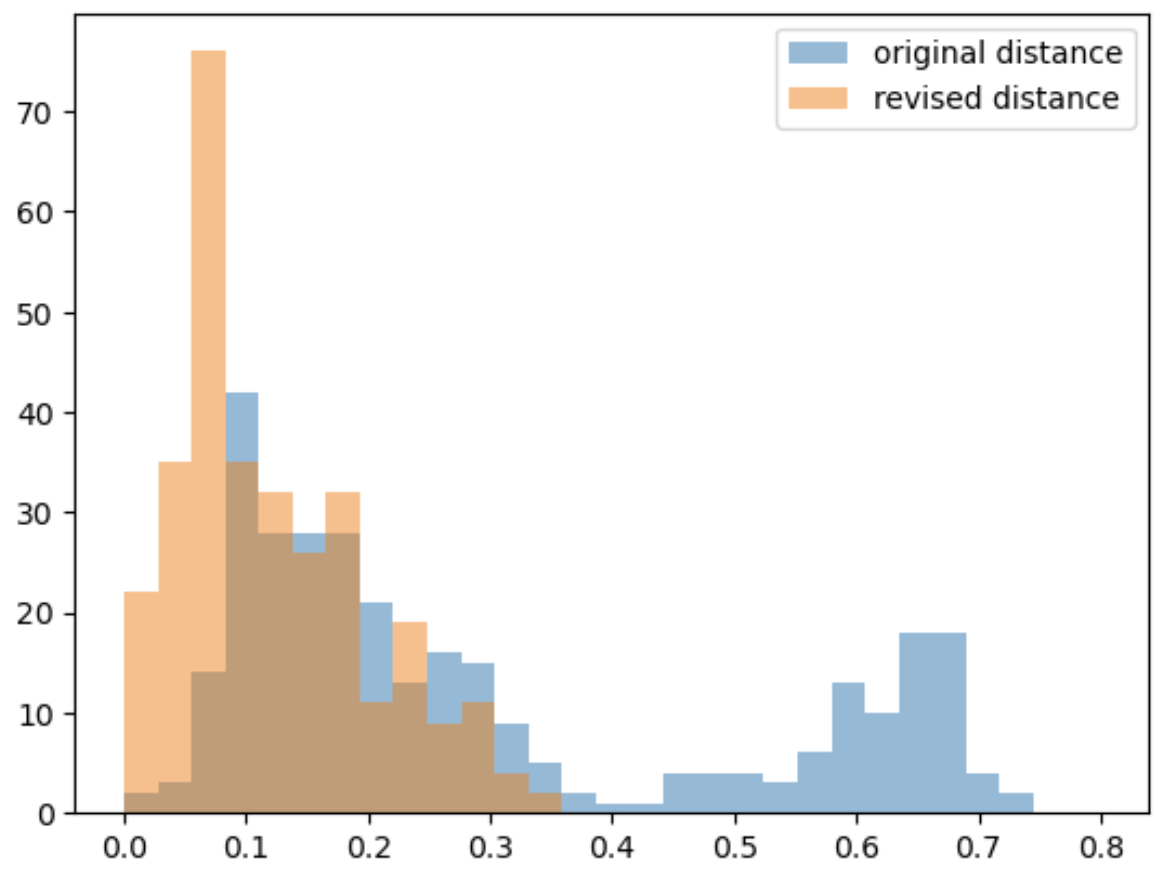} 
    \caption{For the data points such that $x_1<0$  and $x_3 = 0$, we check how similar the explanations are with respect to $\bm{e}_c$. We visualize this as a histogram with the distance on the horizontal axis. Refining the notion of distance to one that is hypothesis-driven, we remove the ``noise'' introduced by $x_2$ and ascertain that the explanations do cluster in the vicinity of $\bm{e}_c$.}
\label{fig:tabular_exp}
\end{figure}

\textbf{Experiment on tabular data. }
On this illustrative task, the model has an AUC of 1.00. Our experiment was geared towards testing whether the explanations matched our hypothesis ``if $x_{i1} < 0$ and $x_{i3} = 0$, then the SHAP-value for $x_{i1}$ is positive.'' 
By choosing a first data point as a point of reference, we obtained a semantic match AUC of 0.99 and a median distance of 0.20. 
These scores indicate that the distances allow us to distinguish data points where the hypothesis is satisfied from those where it is not, but explanations are not coherent. Inspection of the histogram of distances of the explanations (Figure \ref{fig:tabular_exp}, \textcolor{blue}{blue} histogram) reveals a group of data points for which the distance is rather large. What is at play here is that $x_2$ is confounding our notion of distance: while $x_2$ is irrelevant for the hypothesis we formulated, our naive notion of distance includes the distance on that dimension. This simple example brings into light the fact that hypotheses may be \textit{local}. Revising our notion of distance to only consider dimension $x_1$ and $x_3$, we see a drop in median distance reaching 0.09 (Figure \ref{fig:tabular_exp}, \textcolor{orange}{orange} histogram), while semantic match AUC remains high at 0.92. 
These observations allow us to conclude that we have a reasonable level of semantic match, and thus we are confident that the explanations match our hypothesis. 
Note that in this process we have not dissected the model itself, which in principle has remained a black box. 

\begin{figure}[!ht]
\vspace{-1\baselineskip}
  \centering

\subfloat{\includegraphics[width=0.45\textwidth]{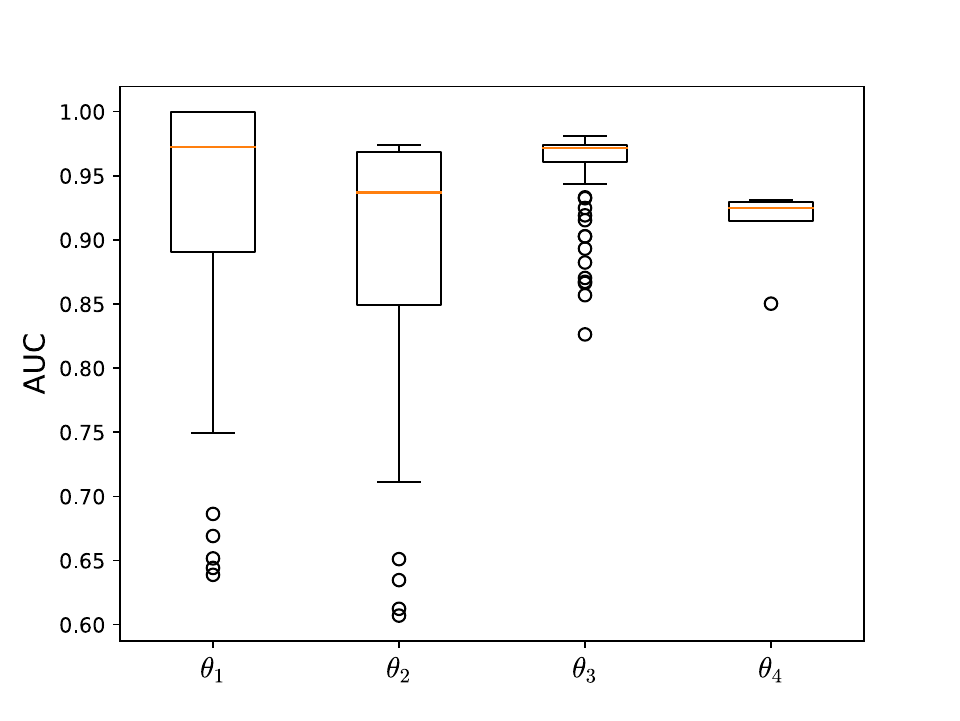}}{}
\subfloat{\includegraphics[width=0.45\textwidth]{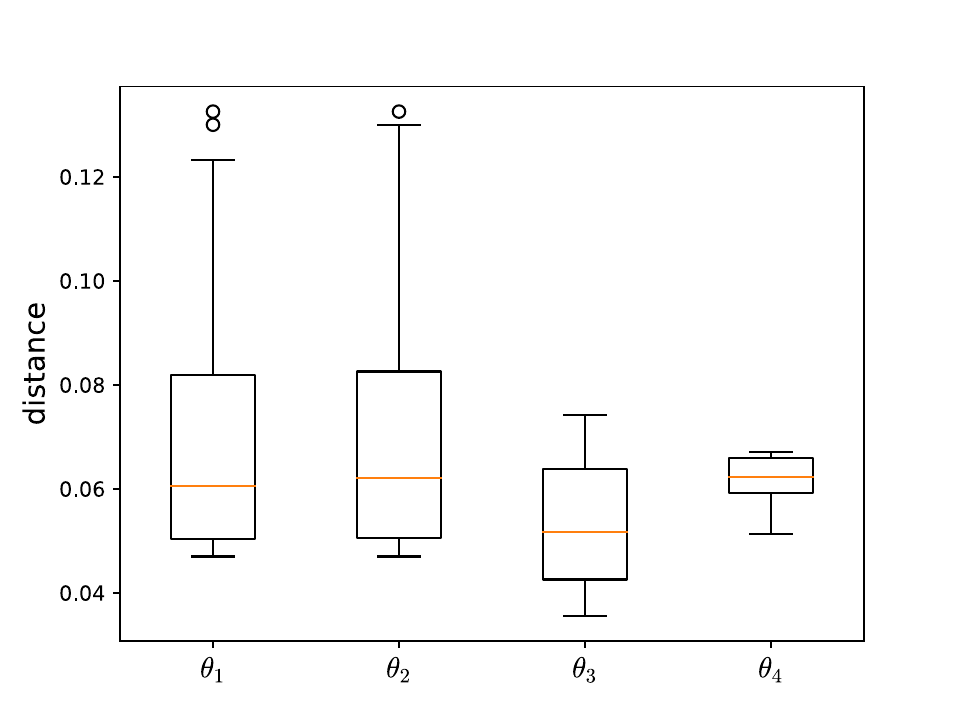}}{}
  \vspace*{-3mm}
  \caption{Boxplots of AUC and median distance for hypotheses based on contribution of the red target object in images from the MALeVIC dataset. The boxplots are obtained by sampling all data points complying with the hypothesis as points of reference.}
  \label{fig:images}
\end{figure}

\textbf{Experiments on MALeViC. }
On the MALeViC classification task (\textit{big} vs. \textit{small}), the model reached an accuracy of 91.5\%. We were thus interested in confirming that the model has learned which shapes it is supposed to consider. We formulated hypotheses with both upper and lower bound for the contribution placed on the red target object, and adjusted for the correctness of the model's predictions:
\begin{itemize}
    \item $\theta_1$: `$\geq 10\%$ of the contribution is placed on the target object'
    \item $\theta_2$: `$\geq 10\%$ of the contribution is placed on the target object and the prediction is correct'
    \item $\theta_3$: `$< 5\%$ of the contribution is placed on the target object and the  prediction is correct'
    \item $\theta_4$: `$< 5\%$ of the contribution is placed on the target object and the prediction is not correct' 
\end{itemize}
Results are summarized in Figure \ref{fig:images}. All hypotheses obtain high AUC, suggesting the explanations clearly separate the data points complying to the hypothesis from the rest. 
Overall, the median distances are relatively small (\textit{i.e.}, the median of median distances stands close to 6\%). This suggests that all hypotheses have good semantic match. In particular, since the lower bound of $\theta_1, \theta_2$ is weak and the reference points for hypotheses $\theta_3, \theta_4$ have less than 5\% of contribution placed on the target objects, this entails that in \emph{all} explanations the contribution of the target object is limited. 
For further insights into the hypotheses considered, we refer to Appendix \ref{appendix:image_data_exp}.

If the focus on the target object does not explain the high performance, perhaps the model is leveraging the smallest and largest objects (these objects determine the label together with the target). Hence we recast the hypotheses, expanding their scope to encompass the contribution of the target, biggest and smallest objects in the image:
\begin{itemize}
    \item $\theta_5$: `$\geq 30\%$ of the contribution is placed on the target, biggest and smallest objects'     
    \item $\theta_6$: `$\geq 30\%$ of the contribution is placed on the target, biggest and smallest objects and the prediction is correct'
    \item $\theta_7$: `$< 15\%$ of the contribution is placed on the target, biggest and smallest objects and the  prediction is correct'
    \item $\theta_8$: `$< 15\%$ of the contribution is placed on the target, biggest and smallest object and the prediction is not correct' 
\end{itemize}
The results, summarized in Figure \ref{fig:images_all} in Appendix \ref{appendix:image_data_exp}, are similar to the ones observed in the first four hypotheses. Hypotheses $\theta_5$ to $\theta_8$ also obtain high AUC (albeit the first two with high variability), suggesting that explanations complying with the hypothesis and those which do not can be separated easily. 
The median distances are small, roughly between 10\% and 20\%, although slightly larger than in the previous set of hypotheses. Overall, the contribution devoted to the biggest, smallest and target object tends to be less than half. 
Since the target, smallest, and largest objects should be the \textit{only} objects affecting the classification, these results prove that the model does \textit{not} behave as desired and relies on spurious correlations. 

\textbf{Experiments on VOC2006. }The model achieved an accuracy of 87.6\% on the classification task. The hypotheses here mimic the first four hypotheses investigated in the MALeViC case but now varying the contribution cutoff. Results for hypotheses $\theta_1$ and $\theta_3$ are summarized in Figure \ref{fig:theta_evolution_1} and for hypotheses $\theta_2$ and $\theta_4$ in Figure \ref{fig:theta_evolution} in Appendix \ref{appendix:image_data_exp_voc}. Notice that at cutoff 90\% the first hypothesis has a median distance around 0.45, and the same occurs for the third hypothesis with cutoff 10\%. The latter is much higher than what was recorded in the MALeViC data for the same hypothesis. These observations suggest that the bulk of data points assign a substantial amount of contribution to the bounding box of the target object, hence the model focuses substantially on the objects to classify,  as desired (contrary to the MALeViC case, here the context plays no role). For further details and results we refer to Appendix \ref{appendix:image_data_exp_voc}.




\begin{figure*}[!htb]
\centering
\includegraphics[width=0.45\textwidth]{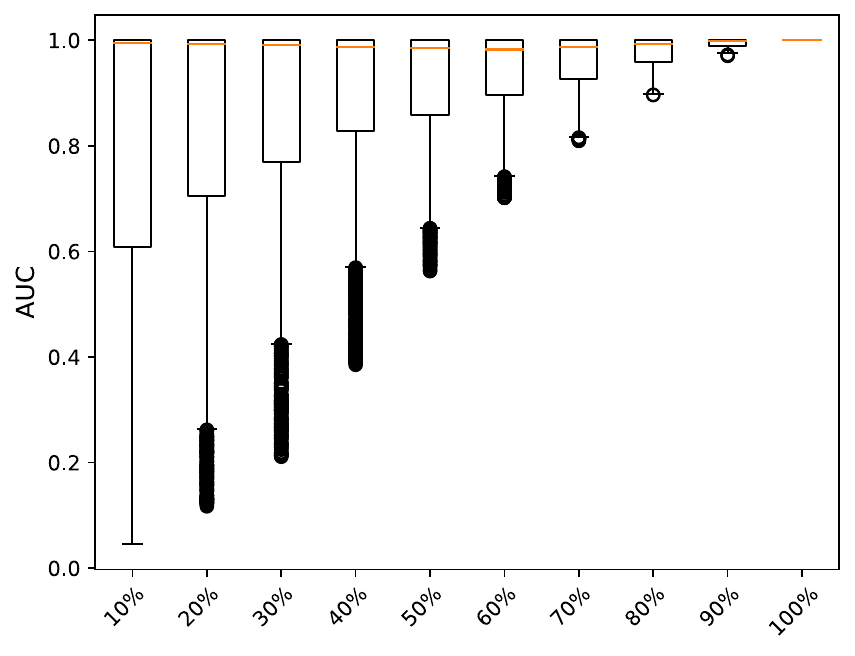}
\includegraphics[width=0.45\textwidth]{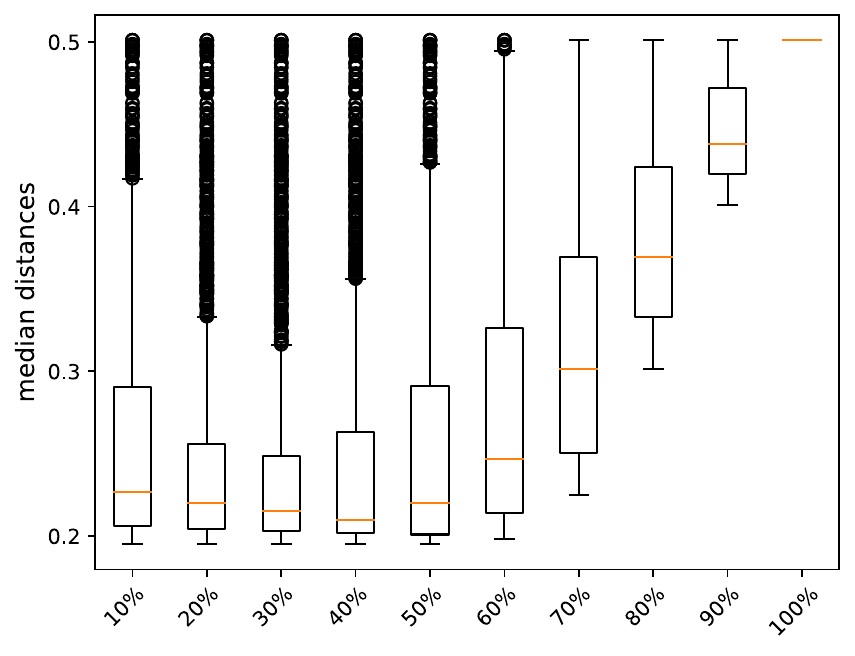}
\includegraphics[width=0.45\textwidth]{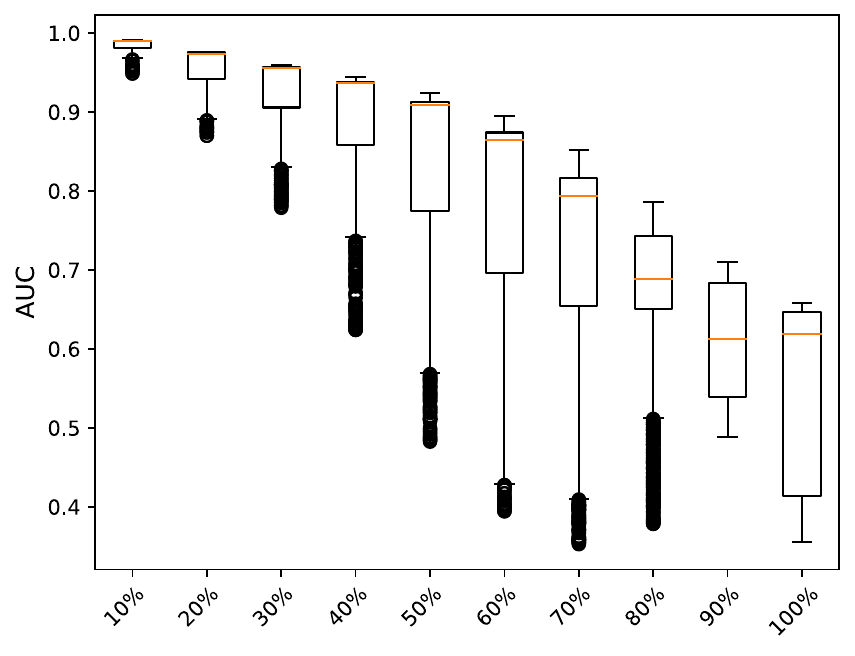}
\includegraphics[width=0.45\textwidth]{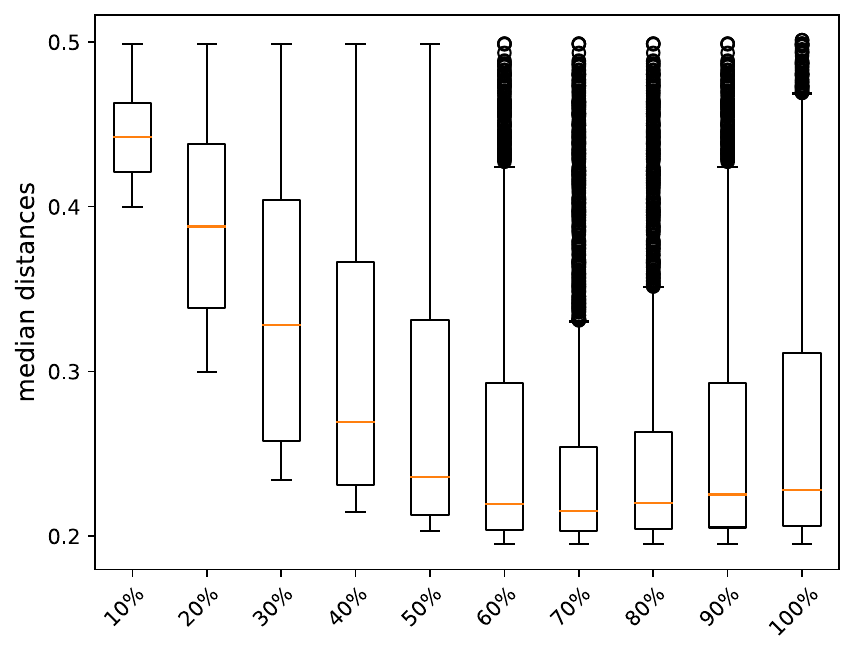}
\caption{AUC and median distance for hypotheses $\theta_1$ (\textit{top}) and $\theta_3$ (\textit{bottom}) with and different thresholds for the VOC2006 dataset.}
  \label{fig:theta_evolution_1}
\end{figure*}

\textbf{Experiments on SQuAD. }
To distinguish the biased from the unbiased model, we consider the following hypotheses:
\begin{itemize}
    \item $\theta_1$: `the contribution of the first sentence is $\geq z\%$ of the total contribution'
    \item $\theta_2$: `the contribution of the first sentence is $\geq z\%$ of the total contribution, the answer is not in the first sentence, and the prediction is wrong (\textit{i.e.}, exact match between model's prediction and correct answer is 0)'
    \item $\theta_3$: `the contribution of the first sentence is $\geq z\%$ of the total contribution, the answer is in the first sentence, and the prediction is correct.
\end{itemize}
Each hypothesis is evaluated with $z\in\{0,10,20,\dots,100\}$. Compliance to $\theta_1$ indicates a lower bound for the contribution of the first sentence. For high thresholds $z$, semantic match wit $\theta_2$ indicates that the model consistently predicts based on the first sentence, when it should rely instead on some other sentence. 
$\theta_3$ is used to check that the unbiased model has learned to rely on the first sentence whenever the correct answer is there.

\begin{figure*}[!htb]
    \centering
    \includegraphics[width=0.45\textwidth]{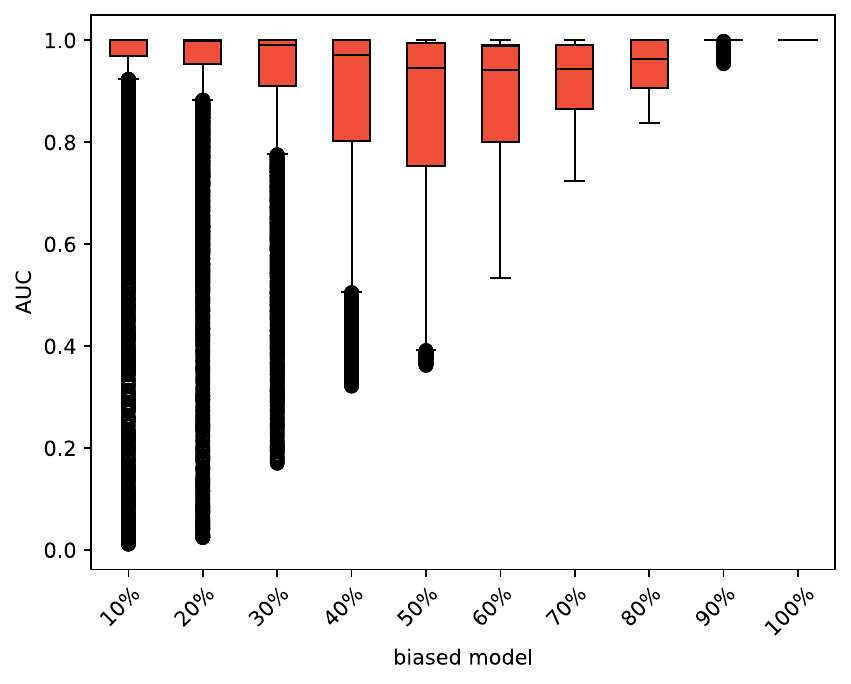} 
    \includegraphics[width=0.45\textwidth]{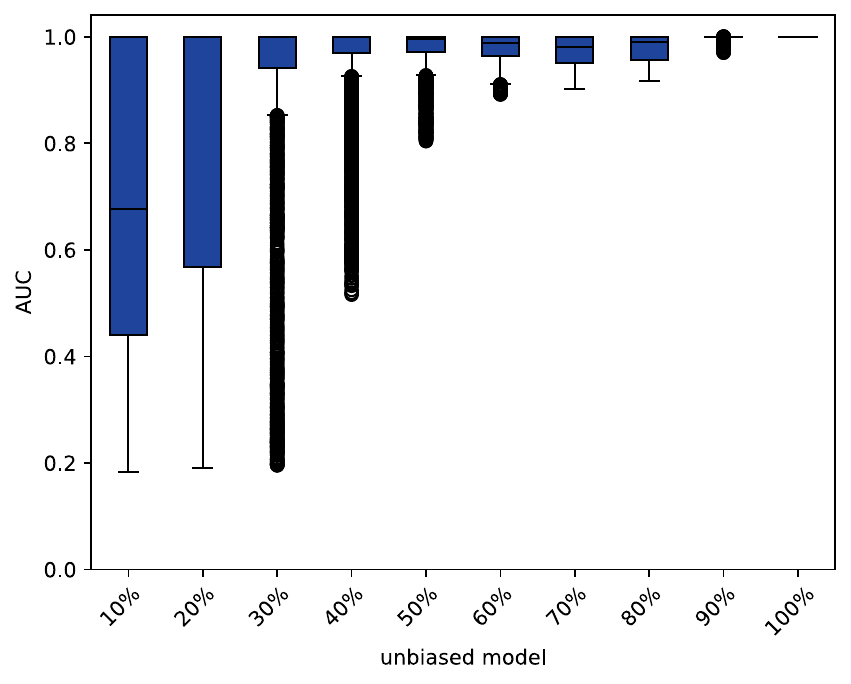} 
    \includegraphics[width=0.45\textwidth]{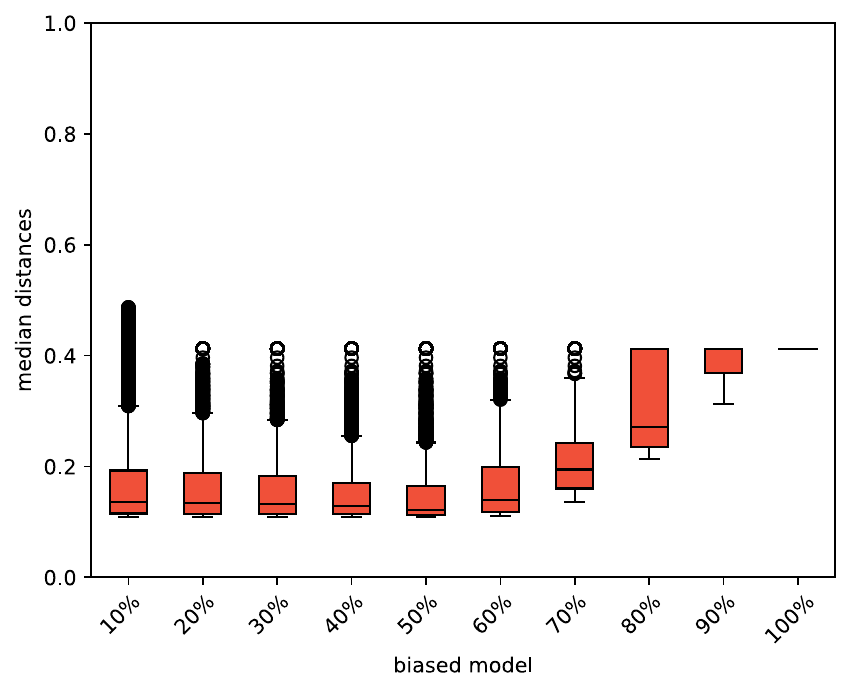}
    \includegraphics[width=0.45\textwidth]{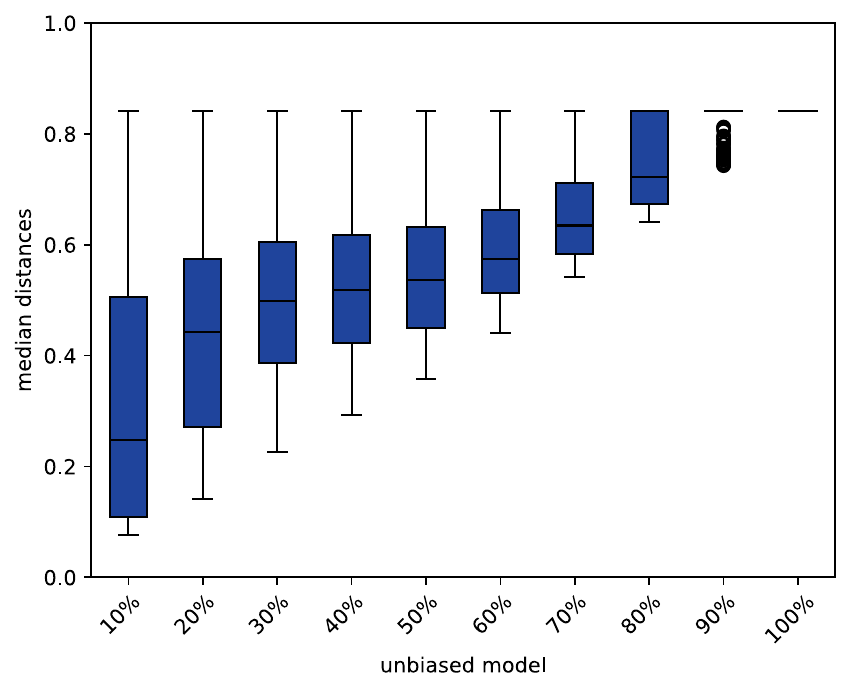}
    \caption{AUC and median distance of biased (\textcolor{b_color}{red}) and unbiased (\textcolor{u_color}{blue}) models for $\theta_1$ and different thresholds on the SQuAD dataset.}
\label{fig:squad_t1main}
\end{figure*}

For $\theta_1$ 
the median distance of the biased model (Figure \ref{fig:squad_t1main}, bottom left) is stable and small across all choices of $z$, reaching its minimum at $z=50\%$, after which it increases to $0.4$. This signals that the majority of data points have a similar behavior (low median distance) having high contribution from the first sentence (median distance stays low as $z$ increases). On the other hand, the unbiased model's median distance plot show an increasing trend, with far higher variance and wider quartiles (Figure \ref{fig:squad_t1main}, bottom right). This is a sign that the contribution of the first sentence is not consistent for the unbiased model, hinting that the model bases its predictions on the first sentence only when the answer is there.
The AUC plots (Figure \ref{fig:squad_t1main}) show that the AUC for the biased model drops at a $50\%$ threshold, indicating that there are many data points where the contribution from the first sentence is around half of the total; those are therefore very close to points that comply with $\theta_1$. On the other hand, the unbiased model's AUC is low at $x=10\%$, suggesting that in most of the data points the contribution of the first sentence is around $10\%$ of the total. The results for the second and third hypothesis (displayed in Appendix \ref{appendix:squad_exp}) are consistent with the above discussion. 
In conclusion, we can say that testing for semantic match we can tell apart the biased from the unbiased model, \textit{e.g.}, explanations of the biased model show great coherence with both $\theta_1$ and $\theta_2$ at $z=50\%$, while the same is not true for the unbiased model. 

\section{Discussion}
In the previous sections, we laid out a procedure to investigate the semantic match between human-understandable concepts and attribution-based explanations. The procedure begins by making the hypothesis more precise, and by defining a notion of distance between explanations that is hypothesis-driven. We then proposed some diagnostic tools to measure the level of semantic match between said hypothesis and the explanations to ground our intuitions and prevent confirmation bias.
This framework was put to the test on one synthetic tabular data, on two computer vision tasks, and on one question-answering NLP task,  eliciting reliable information on model behavior.

As for limitations, we note that this whole endeavor hinges on the assumption that explanations have some degree of faithfulness to the model~\citep{jacovi-goldberg-2020-towards}. If explanations misrepresent the model, a semantic match is not going to give us reliable information about the model. Moreover, we only experimented with SHAP, which is widely used but only one of the many feature attribution techniques. 

The experiments revealed some interesting aspects of semantic match. First, often hypotheses only pertain to a part of the input sample, and it may not be straightforward to define a relevant notion of distance (see the bounding box problem in Section \ref{sec:setup}). Second,  results can be sensitive to the specification of the hypothesis, highlighting the importance of formalizing hypotheses precisely and testing different options. Third, sharpening the hypothesis does not necessarily lead to crisper results, see for instance $\theta_1$ and $\theta_2$ from the MALeVIC experiment. Finally, the role of the logical structure of the hypotheses remains to be investigated. 

In future work, we plan on tackling more real-world tasks and hypotheses types, as well as expand to other feature attribution methods. We also intend to go beyond the manual specification of hypotheses and automate the search for matching hypotheses, moving XAI away from guesswork and towards reliable behaviour discovery.

\newpage



\bibliography{main}

\newpage
\appendix
\onecolumn

\section{Details and additional results on the MALeViC experiments}\label{appendix:image_data_exp}

We first explain how the outcome (big vs small) is determined when the data is generated. For each image, a threshold $T$ is computed as follows: $T(I) = Max - k(Max-Min)$, where $I$ is the image, $k$ is randomly sampled from the normal distribution of values centered on $0.29$ ($\mu = 0.29$, $\sigma = 0.066$), 
and $Max$ and $Min$ are the areas, in pixels, of the biggest and smallest objects in $I$, respectively. During the construction of the dataset, an object is deemed \textit{big}, if its area exceeds $T$; otherwise, \textit{small}. In our experiments we only consider the label of the (unique) red shape in the image.

\begin{figure*}[htbp!]
    \centering
    \includegraphics[width=0.85\textwidth]{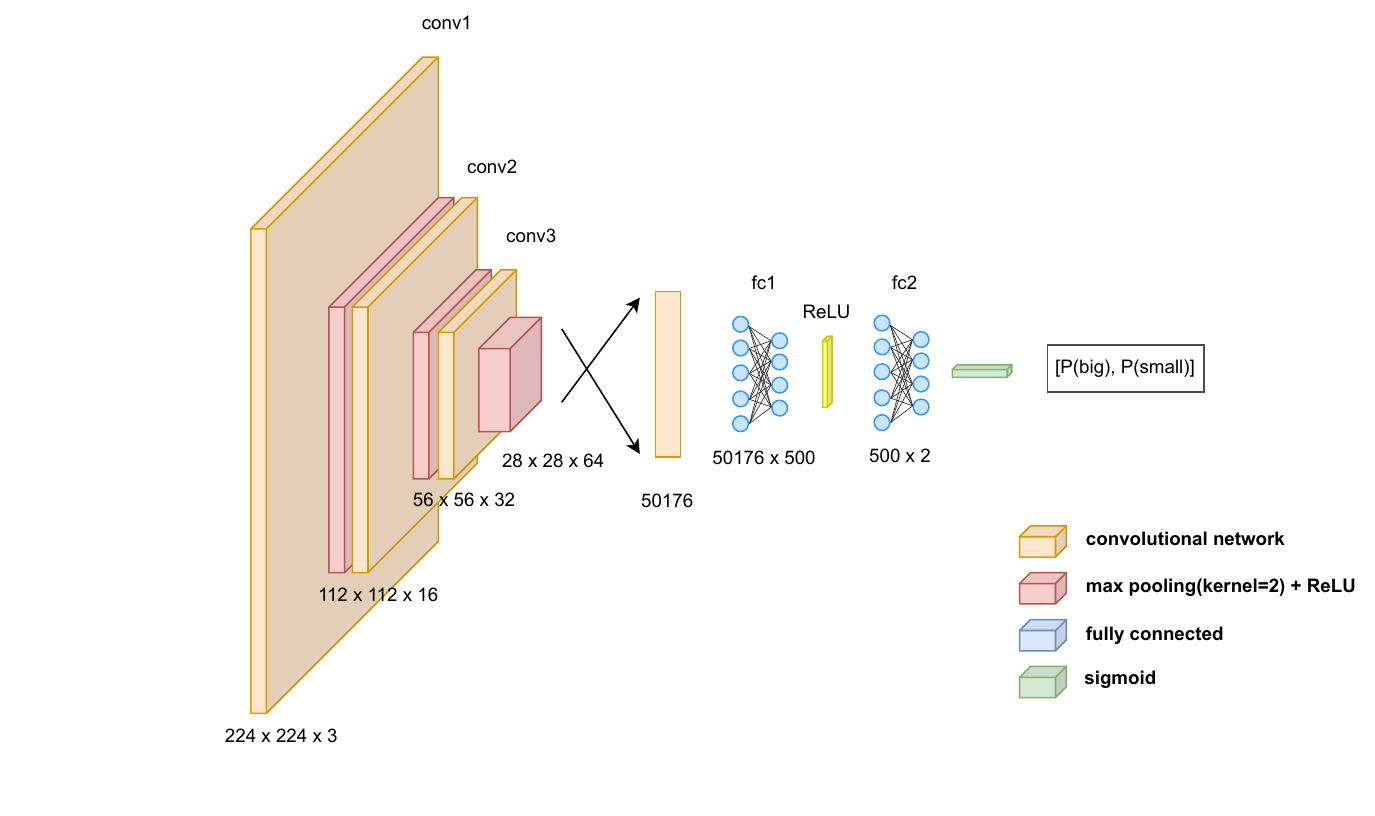}
    {\caption{CNN Architecture}
    \label{fig:architecture}}
\end{figure*}

Next, we describe the CNN architecture used for the experiments. The input for the convolutional neural network is a 3-channel image containing squares or rectangles. The model used consists of three convolutional layers with 3,16 and 32 filters respectively. The dimension of the filters is 3x3. Each convolutional layer is followed by max pooling (2x2 filter) and a rectified linear unit (ReLU) activation function. The output is flattened and passed through a dropout layer (25\% rate), two fully connected layers and a sigmoid to output probabilities. The overall architecture is shown in Figure \ref{fig:architecture}.
The network was trained for 20 epochs using a batch size of 128. The chosen optimization algorithm was Adam with a learning rate set to 0.001 using the binary cross entropy loss. The random seed is set to 42. During training, the model with lowest validation loss was selected for inference.

\begin{figure}[htp!]
  \centering
 \includegraphics[width=0.45\columnwidth]{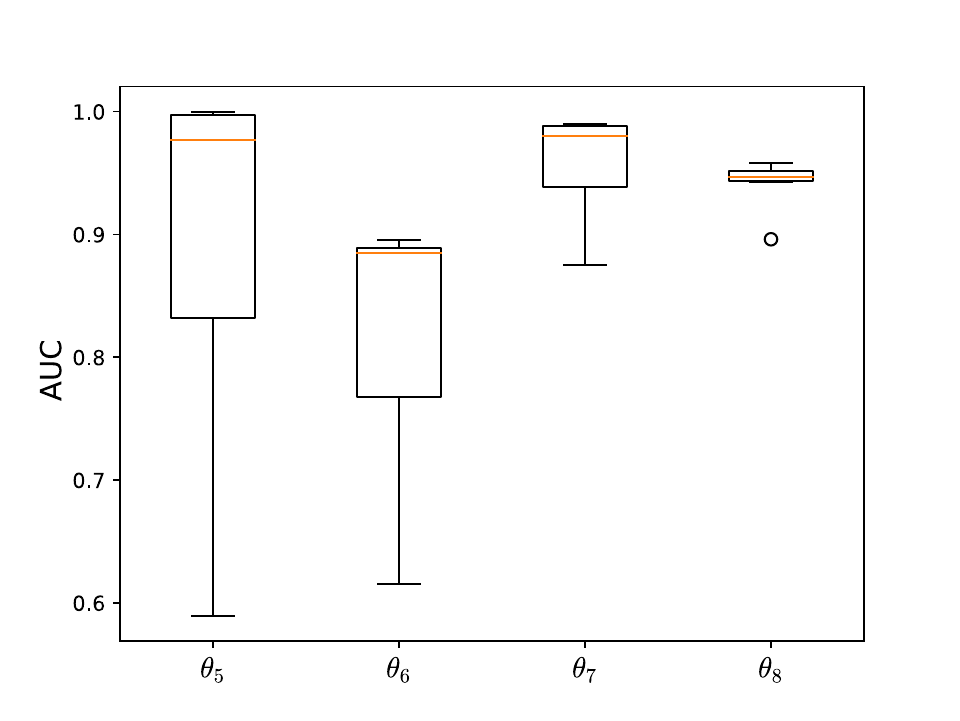}
 \includegraphics[width=.45\columnwidth]{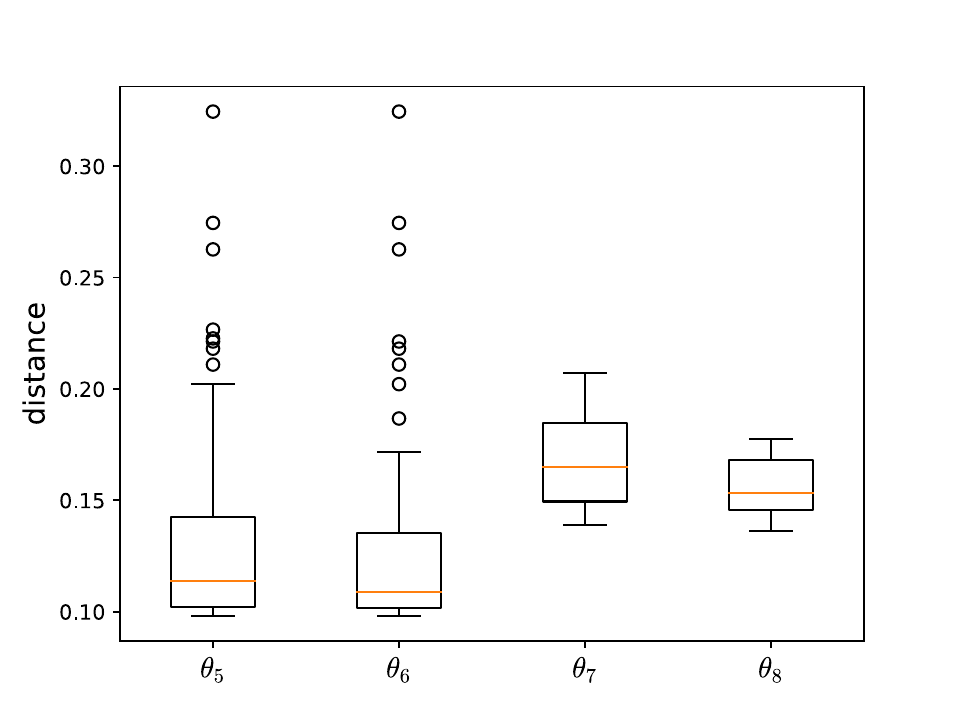}
  \caption{Boxplots of AUC and median distance to assess semantic match for hypotheses based on contribution of the red target, biggest and smallest objects in images from the MALeVIC dataset. The boxplots are obtained by sampling all data points complying with the hypothesis as points of reference.}
  \label{fig:images_all}
\end{figure}

Finally, we expand on the result presented in the main text. In Figure \ref{fig:images_all} we display the boxplots of AUC and median distance to assess semantic match for hypotheses based on contribution of the red target, biggest
and smallest objects, which were mentioned in Section \ref{sec:results}.

These experiments were conducted using an Ubuntu 20.04 machine with an A10 GPU. Python v3.11.4 was used for the  experiments overall, with OpenCV package version 4.8.0.76 for image processing, scikit-learn version 1.3.0 for metrics calculation and shap package version 0.42.1 for SHAP values estimation. PyTorch v2.0.1 and torchvision v0.15.2 were used for implementing CNN frameworks. The completion of the MALeViC experiments took approximately 2 days (including model training, mask generation and SHAP value calculation).

\section{Further details and results on the PASCAL Visual Object Classes (VOC2006) experiments}
\label{appendix:image_data_exp_voc}
\begin{figure*}[!htb]
    \centering    \includegraphics[width=0.95\textwidth]{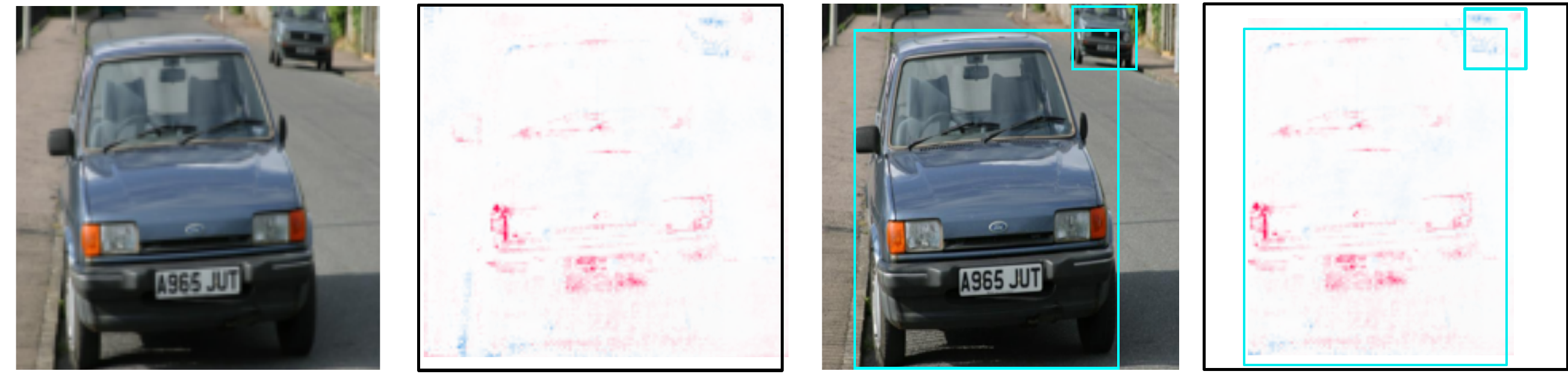}
    {\caption{An example image from the VOC2006 dataset (\textit{left}), alongside the SHAP values generated by the model (\textit{center-left}), the \textcolor{cyan}{cyan} bounding box around the object of interest (\textit{center-right}) and the SHAP values after masking is applied (\textit{right}).}
    \label{fig:shap_cars_bbox}}
\end{figure*}

\begin{figure*}[!htb]
    \centering
    \includegraphics[width=0.45\textwidth]{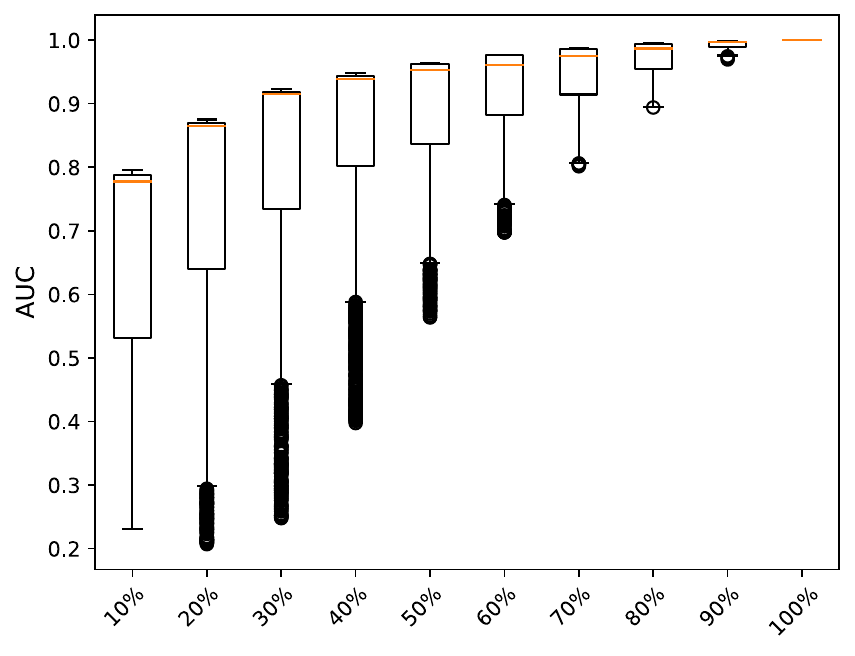}
    \includegraphics[width=0.45\textwidth]{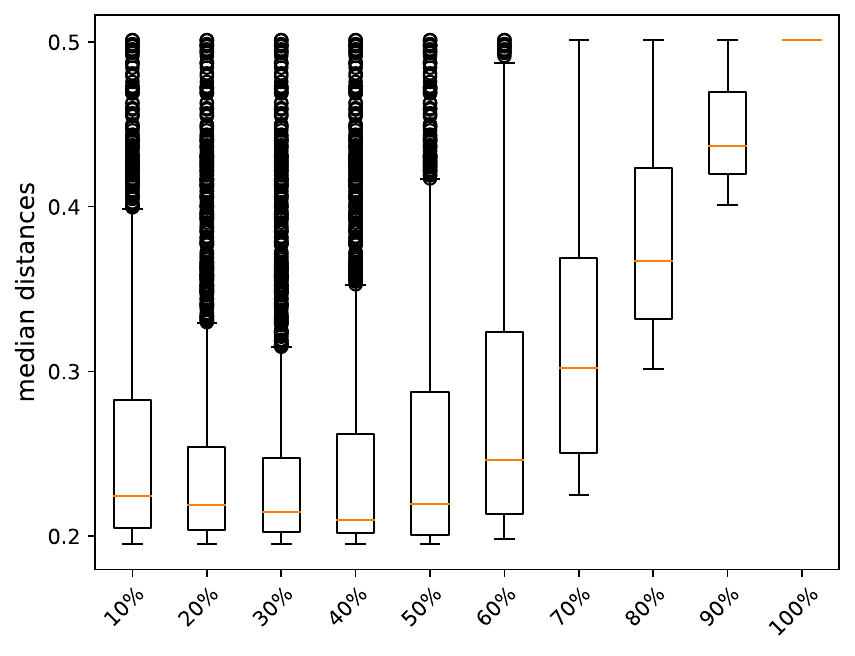}
    \includegraphics[width=0.45\textwidth]{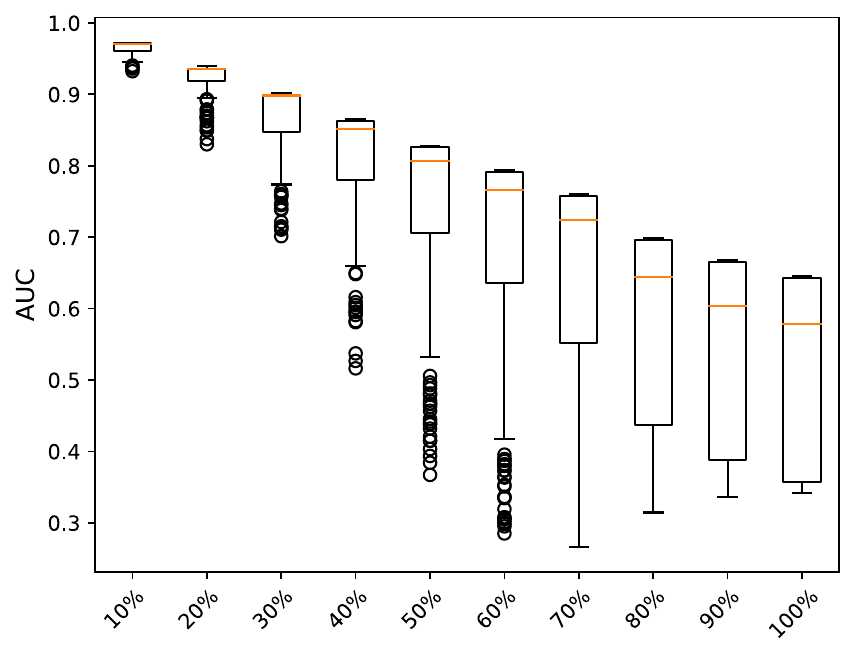}
    \includegraphics[width=0.45\textwidth]{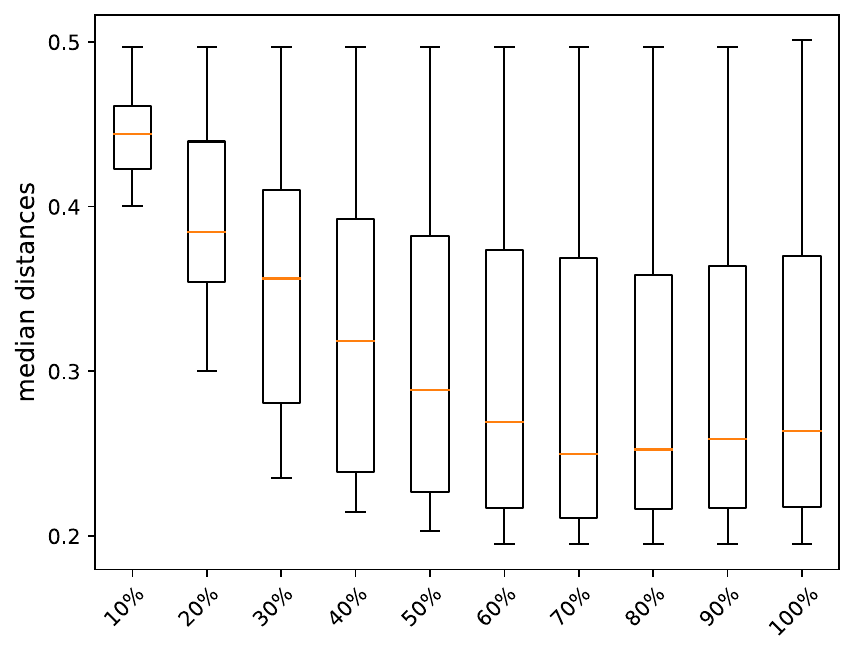}
    \caption{AUC (\textit{left}) and median distances (\textit{right}) for hypotheses $\theta_2$ and $\theta_4$ with increasing contribution thresholds on the target object for the VOC2006 dataset.}
  \label{fig:theta_evolution}
\end{figure*}

As mentioned,  explanations were  generated using pixel-level SHAP values based on the bounding boxes. As done for the MALEVIC dataset, the absolute value of the SHAP values for all bounding boxes was added up and the proportion from the whole image subsequently calculated. This is exemplified in Figure \ref{fig:shap_cars_bbox}. As for additional results, we display in Figure \ref{fig:theta_evolution} the boxplots of AUC and median distance for hypotheses $\theta_2$ and $\theta_4$. 

These experiments were conducted using an Ubuntu 20.04 machine with an A10 GPU. We employed Python v3.11.4 and the same packages used for the MALeViC experiments and the random seed was set to 42 in all experiments, too. The completion of the VOC2006 experiments took approximately 2 days (including model training, mask generation and SHAP value calculation).

\section{Further details and results on the SQuAD experiments}
\label{appendix:squad_exp}

\textbf{Detailed setup. } We obtained biased datasets on the 1st, 2nd, 3rd, 4th and from 5th to last sentence of the context from \citet{ko2021look}, and generate the unbiased dataset by randomly sampling the same number of data points from each of the biased datasets until obtaining approximately the same number of data points of the dataset biased on the first sentence. 
Following \citet{ko2021look} we fine-tuned the BERT-base model (uncased version). The inputs provided to the tokenizer are the question and context of each data point, and the tokenized data is then passed to the model, which tries to predict the tokens of the answer. The model returns two independent probability distributions on each token $t$, the first, $p_{s}$, expresses the probability that a token is the starting token of the answer, while the second, $p_{e}$, expresses the probability that the token is the ending of the prediction. An explicit function computes the score of all possible combinations of starting ($t_s$) and ending ($t_e$) tokens as: $p_{s}(t_s) + p_{e}(t_e)$ and returns the answer with the highest score. The function also selects valid answers (\textit{i.e.}, answers with starting token before ending token), and allows answers with at most 30 tokens.

We train the models for 2 epochs, with Adam optimizer; we set the learning rate at 0.00003. The batch size is 16, the maximum length of the model's input is 384 with stride 128 and truncation only in the context. We set the seed to 42. For the evaluation we use f1-score and exact match, which are standard for SQuAD.

\begin{figure}[!b]
    \centering
    \includegraphics[width=0.75\textwidth]{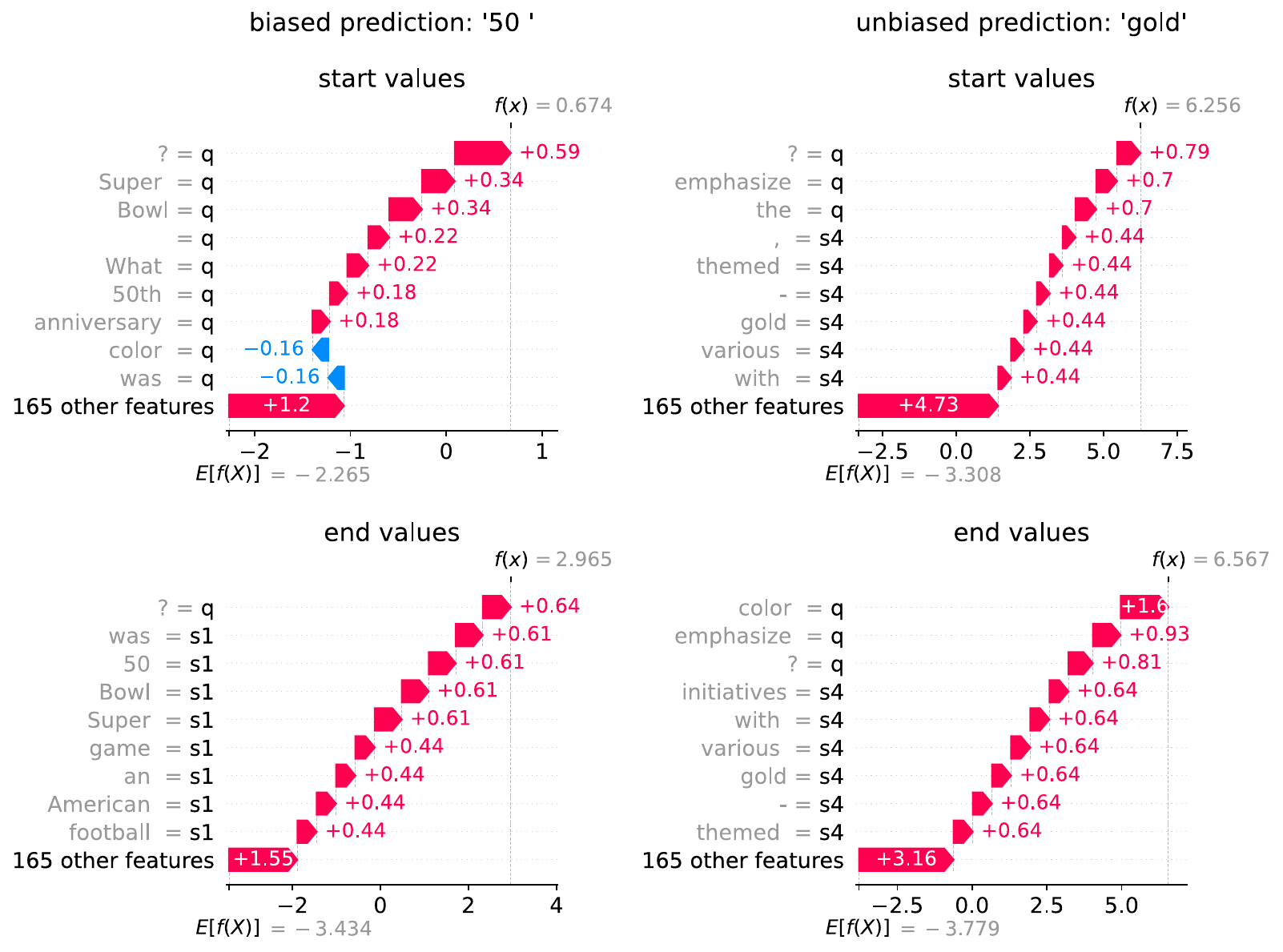} 
    \caption{On the first row, visualisation of starting SHAP values for biased and unbiased models; on the second row visualisation of ending SHAP values for both models. We selected a data point for which both models predict an answer with a single token, as it is simpler to visualize; `50' is the biased model prediction while `gold' is the unbiased model prediction. The final value of the token is the sum of the base value ($E [ \textit{f}(X) ]$) and positive (red) or negative (blue) contributions from all the other tokens in the question and context. On the y-axis the words of the question and context appear in gray, with a black label signaling if each was part of the question (q), the first sentence (s1), the second sentence (s2) and so forth. On the x-axis we plot the SHAP value of each token and the final total.}
\label{fig:squad_shap_visual}
\end{figure}

After training, we compute the SHAP values for the biased and unbiased model on the validation set. 
For each token of the answer we obtain a contribution vector with $n$ entries, where $n$ is the total number of sentences in the context;\footnote{We exclude the contributions from the question, as we expect them to be relevant for the prediction in both models. Re-running the experiment including the question we obtain similar results.} Each entry is the normalized sum of the contributions of each token in that sentence to the final SHAP value of the answer token. We do this both for starting and ending values.
For each sentence (\textit{i.e.}, each entry in the contribution vector) we compute the maximum and average between the starting and ending contribution (in our results we use the average contribution). The model can either return as prediction a single token (as in Figure \ref{fig:squad_shap_visual}) or multiple tokens. In the latter case we perform the above passages for each token of the prediction and then average across the tokens. 

While computing the values, we find a bug in the SHAP library: the masking function used to evaluate the importance of features sometimes malfunctions when proper names appear in the question due to inconsistent tokenization, producing a longer output which in turn throws a length mismatch error. As the number of data points producing this error is around 10\% of the total validation dataset, we exclude them from the experiment, listing their ids in the code repository for reproducibility.\footnote{
github.com/anonymized-for-submission} 
 To generate the mapping between tokens and sentences we use the SpaCy sentencizer of the en\_core\_web\_lg model  \citep{spacy2023}.

\begin{figure}[!htb]
    \centering
    \includegraphics[width=0.45\textwidth]{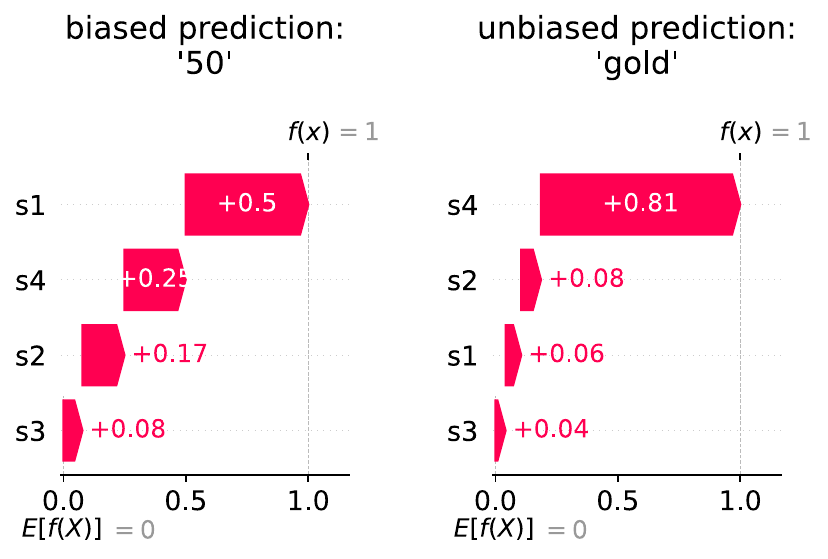}
    \caption{Processed contribution per sentence, for the unbiased model on the right, and for the biased on the left. On the x-axis we plot the proportion of contribution on each sentence, while on the y-axes we plot each sentence. The distance between two data points for hypothesis $\theta_1$ and $\theta_2$ is measured by the absolute difference of the contributions on the first sentence, $d(\boldsymbol{e}_i,\boldsymbol{e}_c)$, with $\boldsymbol{e}_i$ and $\boldsymbol{e}_c$ the explanations of a data point and the reference data point $\boldsymbol{x}_c$, as defined in  Section \ref{sec:methods}. 
    }
\label{fig:squad_att_visual}
\end{figure}

To showcase how we process SHAP values, in Figure \ref{fig:squad_shap_visual} we plot both starting and ending SHAP values of the biased and unbiased models on the same data point, whereas in figure \ref{fig:squad_att_visual} we plot the processed contribution for the same data point. The two models give different predictions, and only the unbiased one is correct in this case.

This experiment was conducted using a ThinkSystem SD650-N v2 node of the Snellius cluster (2 Intel Xeon Platinum 8360Y CPUs and 4 Nvidia A100 GPus). The node was not fully used, with only the heaviest jobs using all the 36 CPU cores and 1 of the 4 GPUs. It took around 1 hour to fine-tune the two models, 2 days for the calculation of SHAP values of all the SQuAD validation dataset and about two hours per hypothesis for the calculation of the metrics, for a total of around 2 days and a half. We used python v3.10.11, pytorch v1.13.0, transformers v4.24.0, tensorboard v2.10.1 and spacy v3.5.3.


\begin{figure}[!b]
    \centering
    \includegraphics[width=0.45\textwidth]{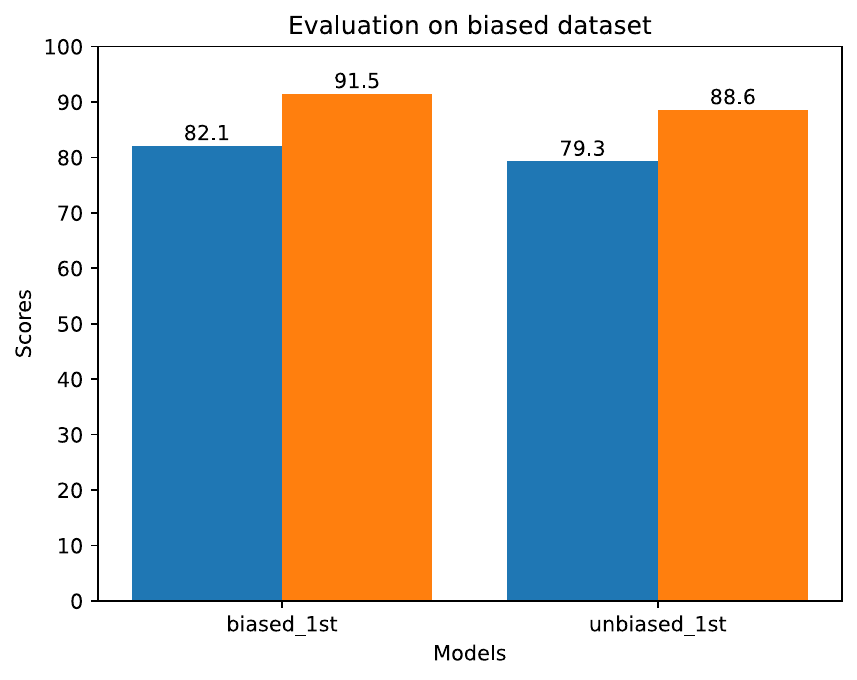}
    \includegraphics[width=0.45\textwidth]{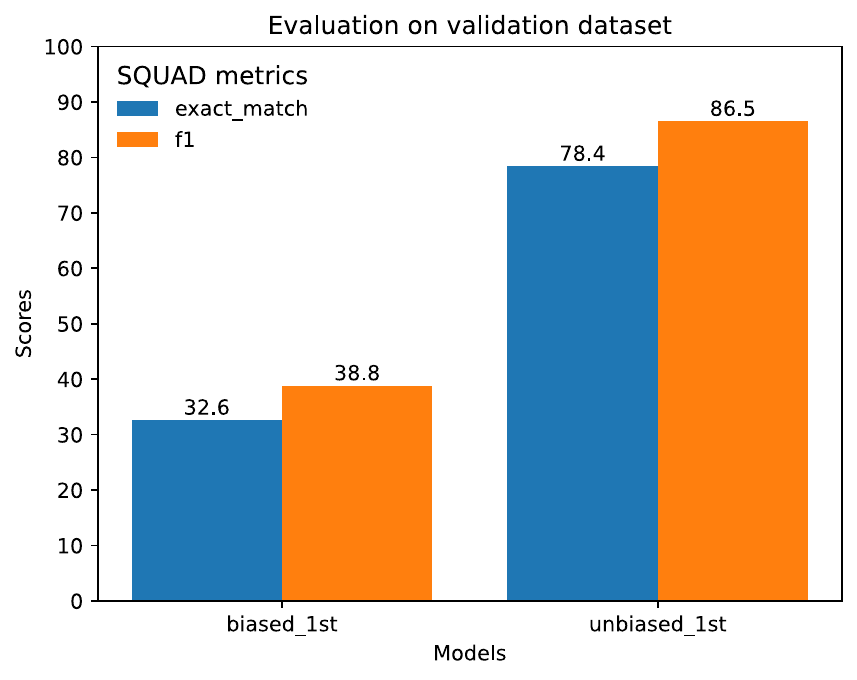} 
    \caption{Performance of biased and unbiased models on the sub-dataset biased on the first sentence and on the standard validation set of the SQuAD dataset. Exact match is displayed in orange, f1 score in blue. The unbiased model performs consistently across the two dataset, while the biased model has good performance only on the biased dataset (which was used to train it).}
\label{fig:squad_eval}
\end{figure}

\textbf{Additional results. }
Evaluating the performance of both models on the biased dataset and on the regular validation set gives us the results in Figure \ref{fig:squad_eval}, which corroborate our initial expectations: the biased model performs better on data points which have the answer in the first sentence, and poorly on the validation dataset, showing that it did not learn to generalize and that there might be a bias; on the other hand, the performance of the model trained on the unbiased subdataset is more consistent (while being only slightly worse on the biased subdataset).

\begin{figure}[!h]
    \centering
    \includegraphics[width=0.48\textwidth]{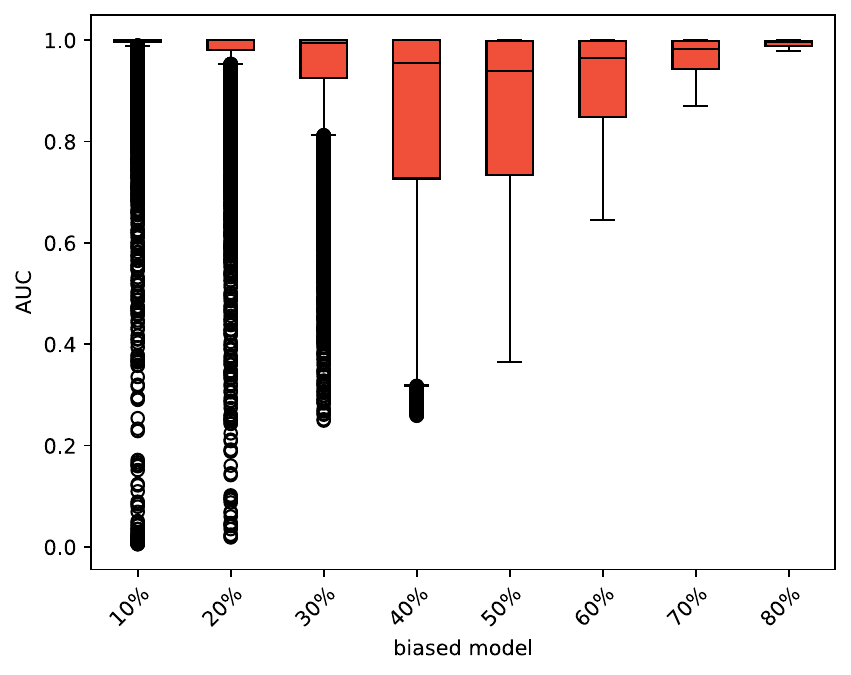} 
    \includegraphics[width=0.48\textwidth]{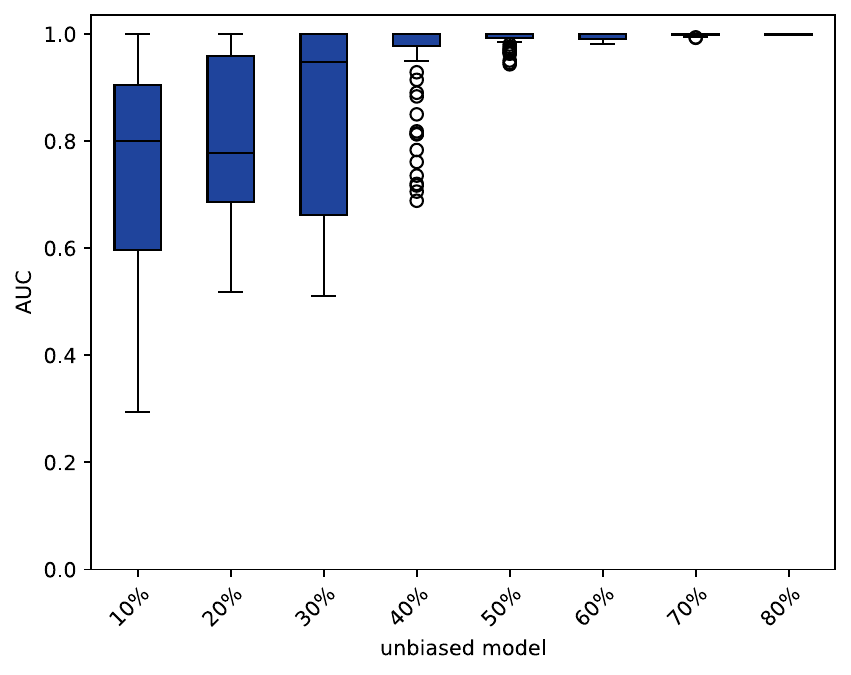} 
    \includegraphics[width=0.48\textwidth]{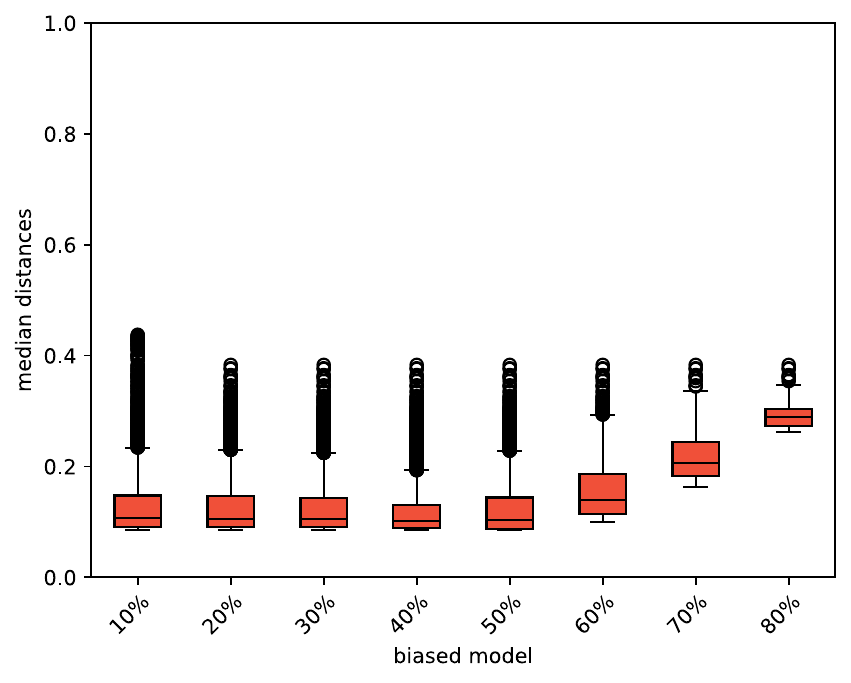}
    \includegraphics[width=0.48\textwidth]{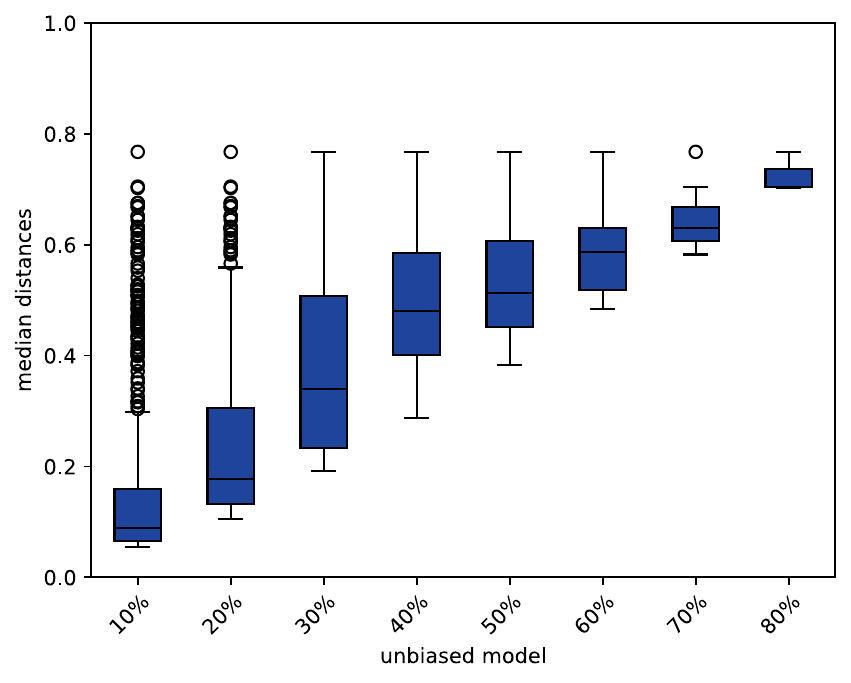}
    \caption{Boxplots of AUC and median distance of biased (\textcolor{b_color}{red}) and unbiased (\textcolor{u_color}{blue}) models for $\theta_2$ and different values of $x$ on the SQuAD validation dataset.}
    \label{fig:squad_t2main}
\end{figure}

We now report the results for the second hypothesis, displayed in Figure \ref{fig:squad_t2main}. Notice that there is no threshold above $80\%$ on the x-axis, showing that no data points satisfy this hypothesis above that threshold. The median distance plots for the biased model have tighter quartiles. As before, there is an almost flat trend, until the minimum is reached, at $50\%$, and then an upward trend begins. The unbiased model shows an almost linear, upward trend, similar to the one for $\theta_1$.
The AUC of the unbiased model also shows a somewhat linear upward trend, which is expected as the higher the threshold, the more the compliant data points are clustered, which in turn makes discrimination easier. On the other hand, the biased model's AUC remains flat and above $0.95$ until threshold $40\%$, where it dips slightly before increasing again. In this respect the plot resembles the corresponding plot for $\theta_1$. As previously, this suggests that, for the biased model, the first sentence has often contribution around $50\%$.  

\begin{figure}[!b]
    \centering
    \centering
    \includegraphics[width=0.48\textwidth]{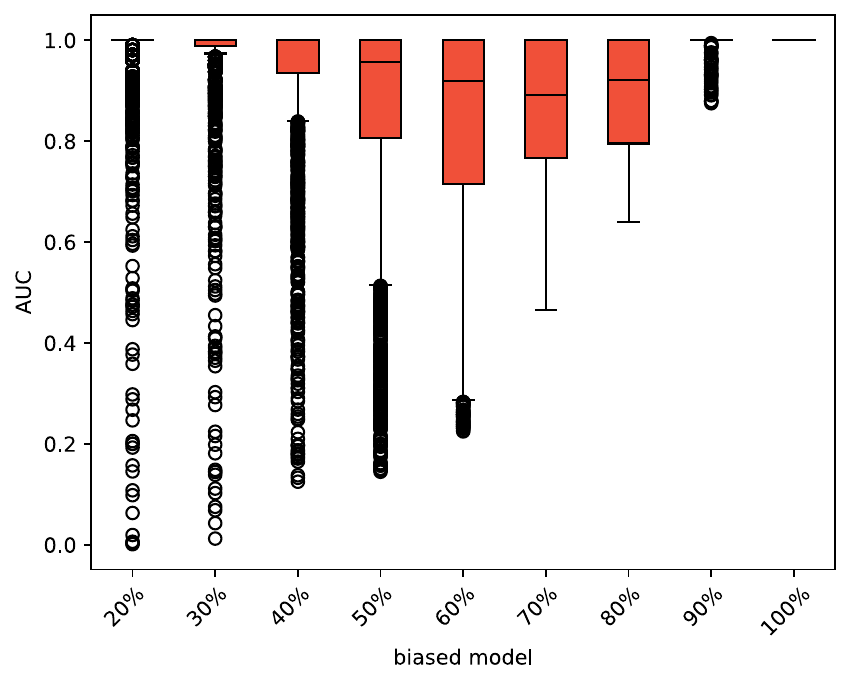} 
    \includegraphics[width=0.48\textwidth]{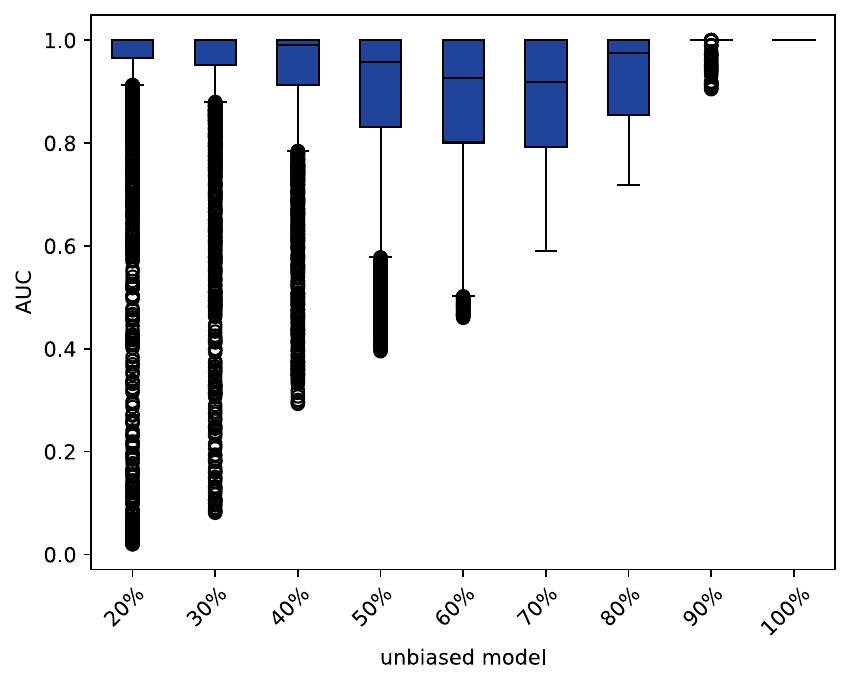} 
    \includegraphics[width=0.48\textwidth]{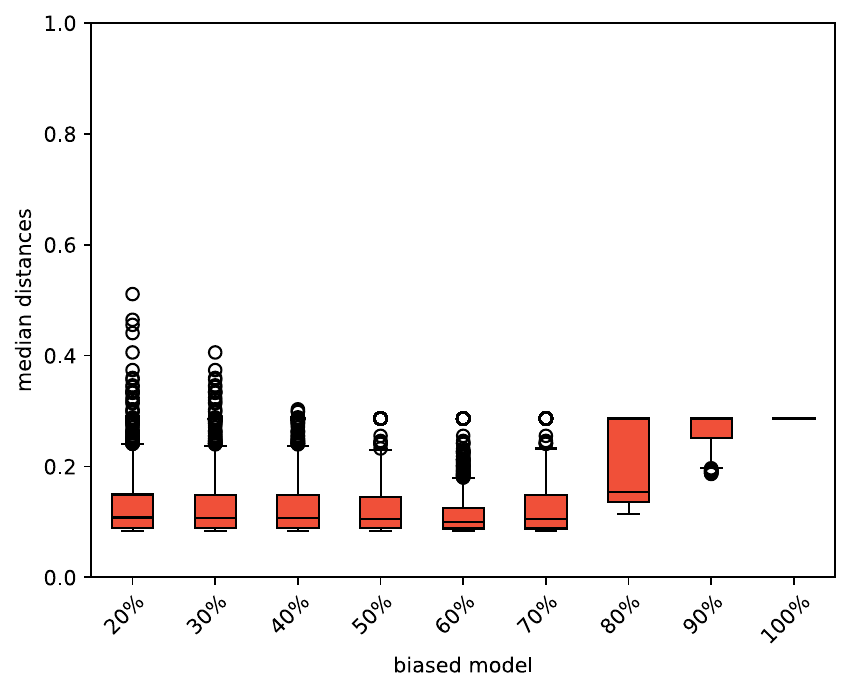}
    \includegraphics[width=0.48\textwidth]{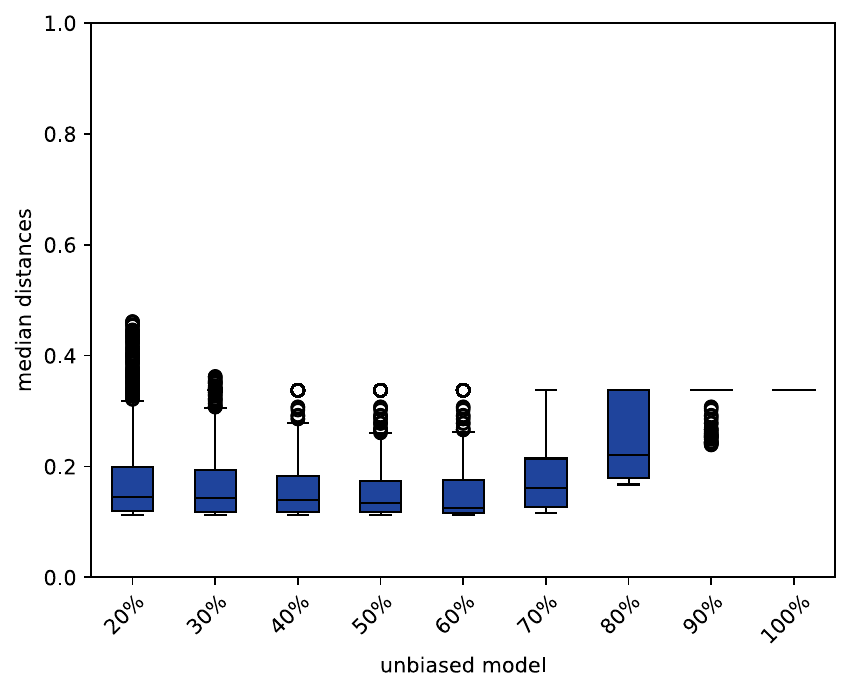}
    \caption{Boxplots of AUC and median distance of biased (\textcolor{b_color}{red}) and unbiased (\textcolor{u_color}{blue}) models for $\theta_3$ and different values of $x$ on the SQuAD validation dataset.}
\label{fig:squad_h3}
\end{figure}

For hypothesis $\theta_3$ we expect something different. This is a `sanity check' hypothesis: we expect both models to put large contribution on the first sentence when the answer is there and they predict correctly. Hence if both pass this check we should see similar plots on both sides. The median distance plot of the third hypothesis (Figure \ref{fig:squad_h3}) is similar to the previous two, essentially confirming that for the biased model the contribution of the first sentence is clustered around 50\%. Contrary to the other hypotheses, now the unbiased model also reproduces the same behavior.   
The AUC plot for the biased model is also similar to the one obtained for $\theta_2$, and again the unbiased model matches it (the numbers are not identical, but the trend is comparable). 
We interpret these results as evidence that the contribution of the first sentence for the unbiased model is high only when the answer is in that sentence. For the biased model, conversely, the first sentence's contribution is high regardless of the position of the true answer.



\begin{figure}[!ht]
    \centering
    \includegraphics[width=0.65\textwidth]{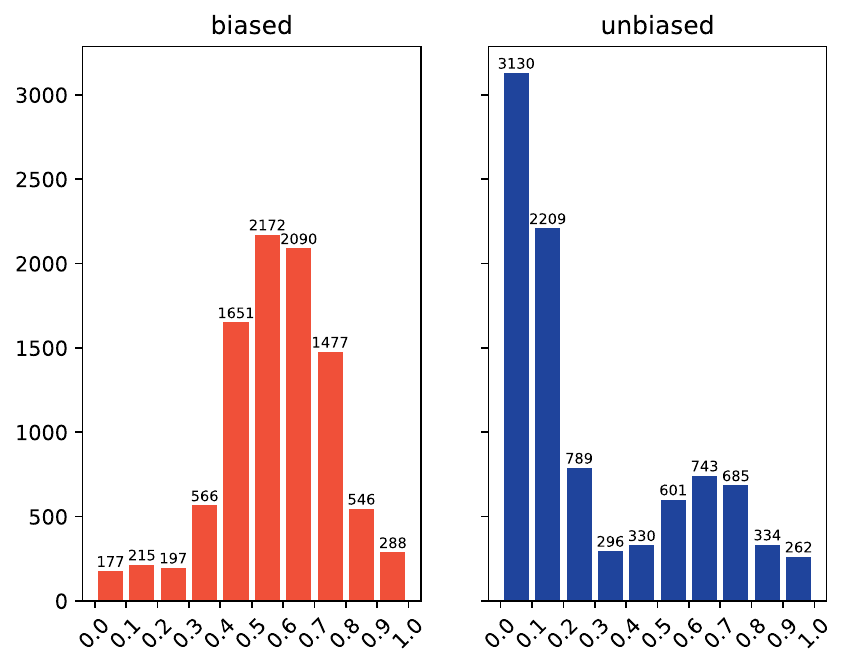}
    \caption{Histogram of the contribution percentage on the first sentence for the biased and unbiased model. Each bar represents a 10\% range of the total contribution.}
\label{fig:squad_contribution_hist}
\end{figure}

\begin{figure}[!b]
    \centering
    \includegraphics[width=0.45\textwidth]{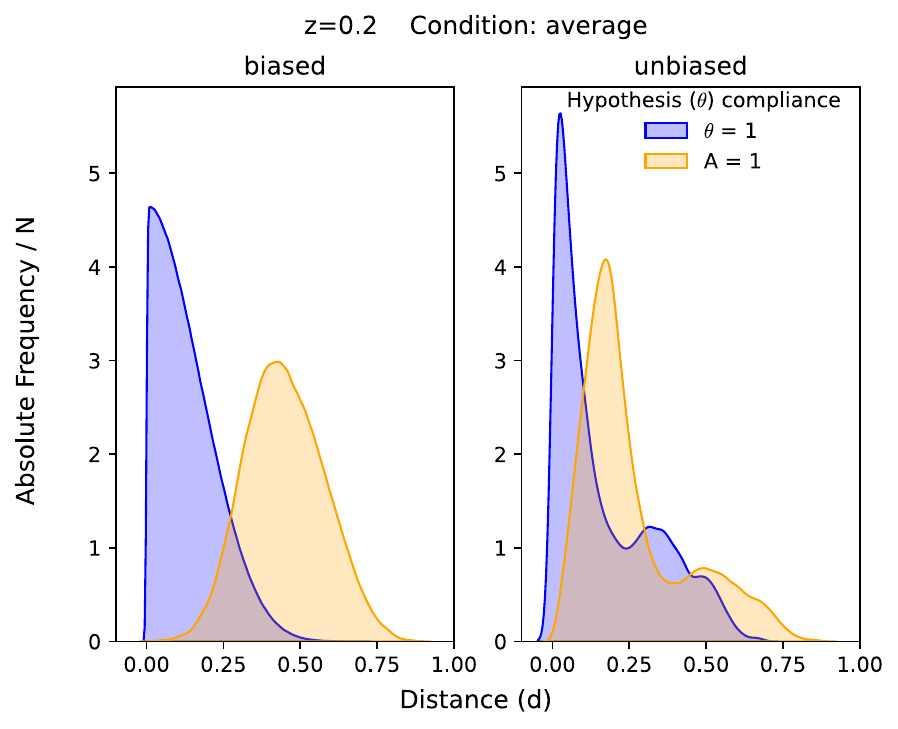} 
    \includegraphics[width=0.45\textwidth]{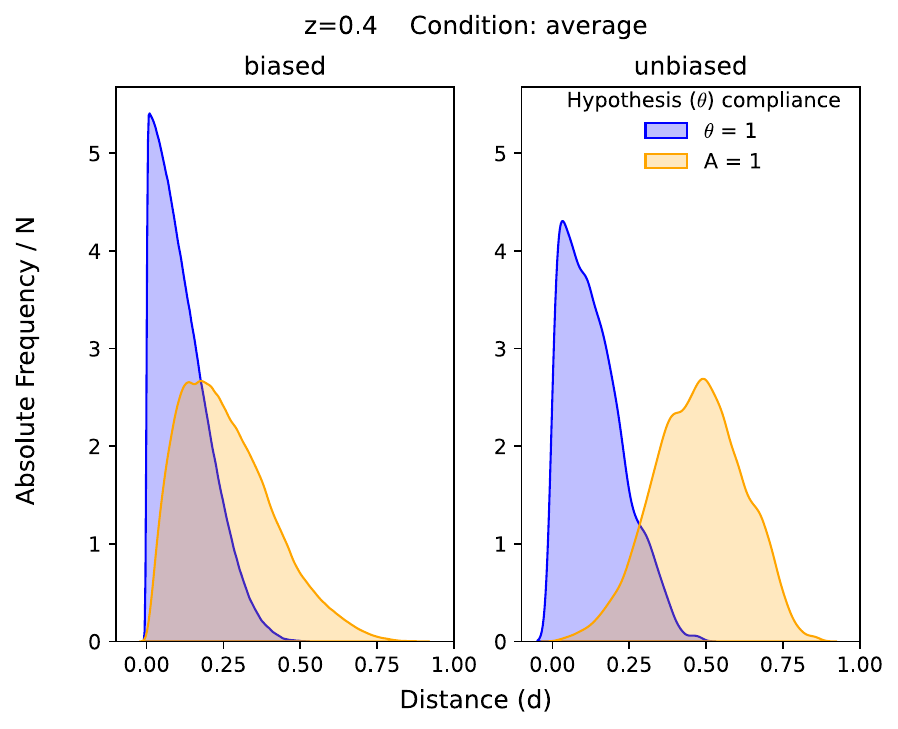}
    \includegraphics[width=0.45\textwidth]{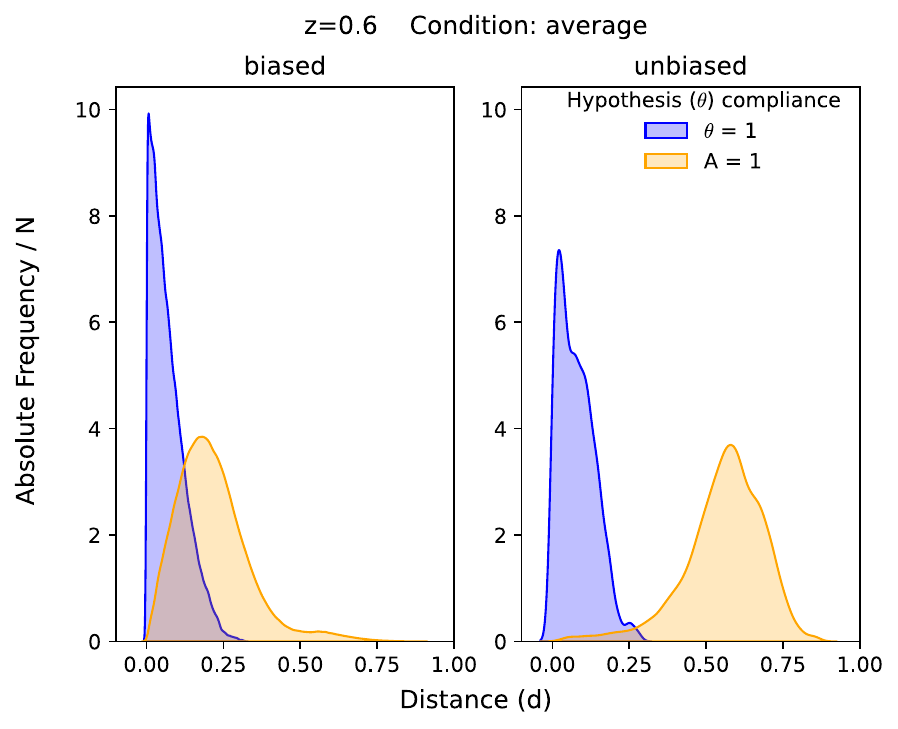}
    \includegraphics[width=0.45\textwidth]{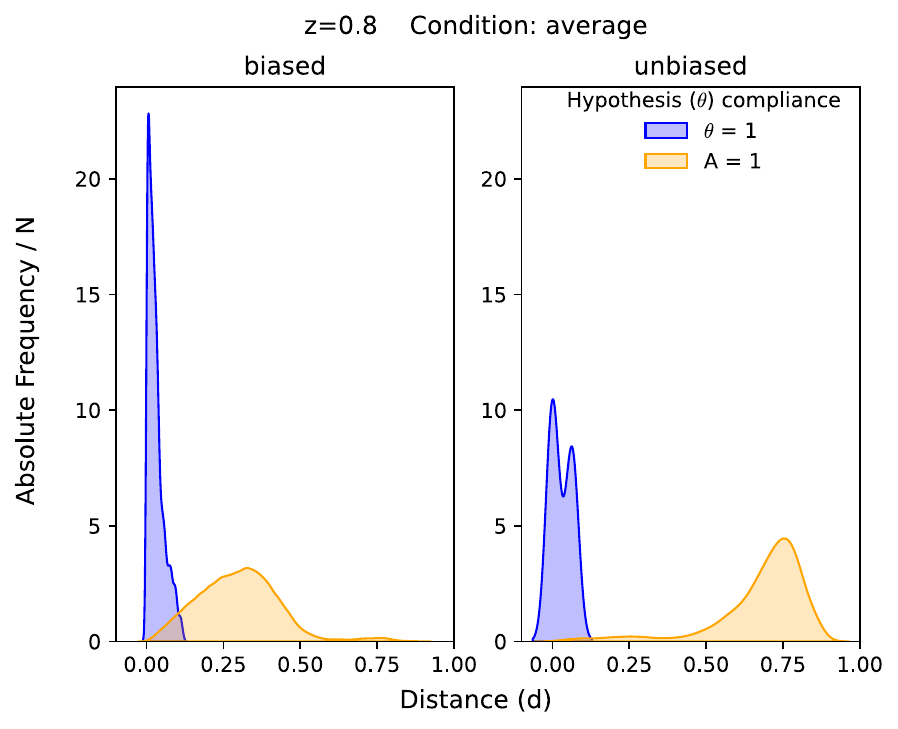}
    \caption{Distance density plots for $\theta_2$ at different thresholds for the biased and unbiased models. The unbiased model (on the right) shows two different peaks for the hypothesis-compliant and non-compliant data points as the threshold increases. The unbiased model, on the other hand, has a lot of overlap between the two sets.}
\label{fig:squad_h2_DD}
\end{figure}

\begin{figure}[!b]
    \centering
    \includegraphics[width=0.45\textwidth]{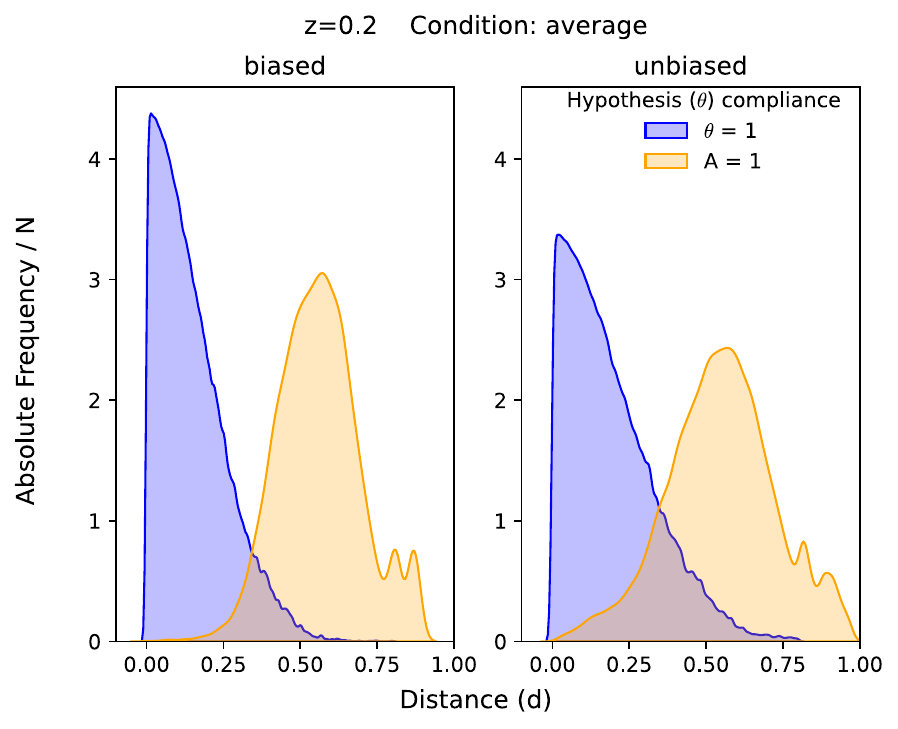} 
    \includegraphics[width=0.45\textwidth]{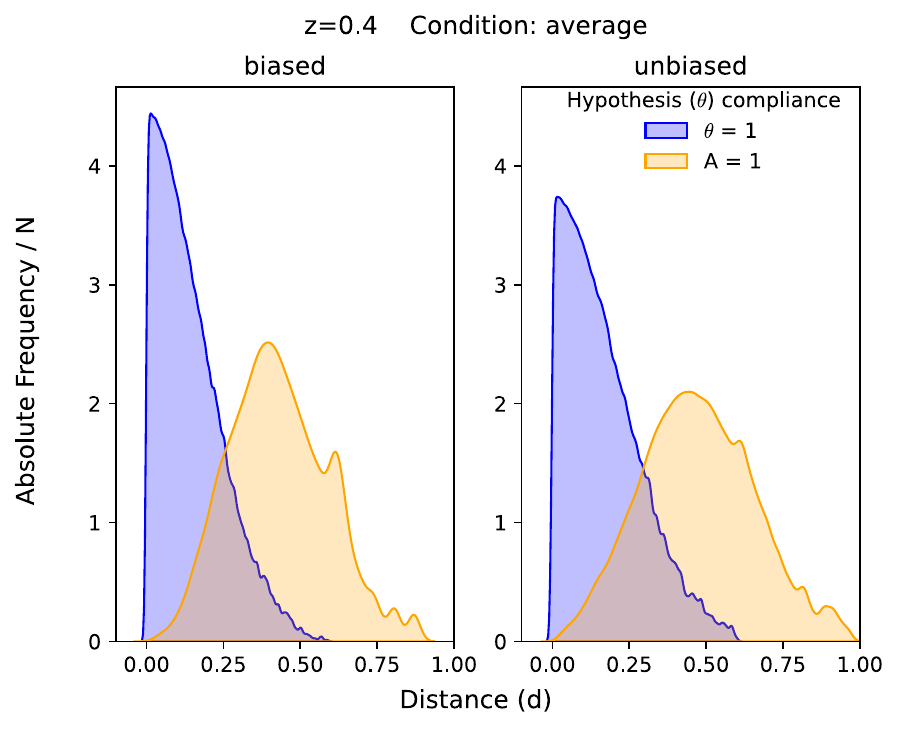}
    \includegraphics[width=0.45\textwidth]{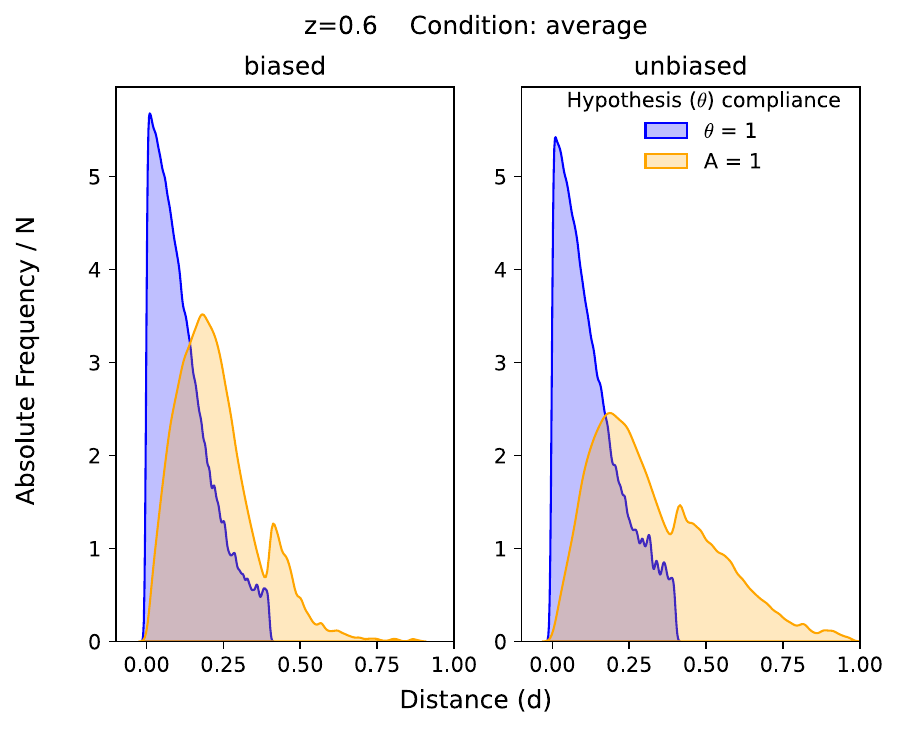}
    \includegraphics[width=0.45\textwidth]{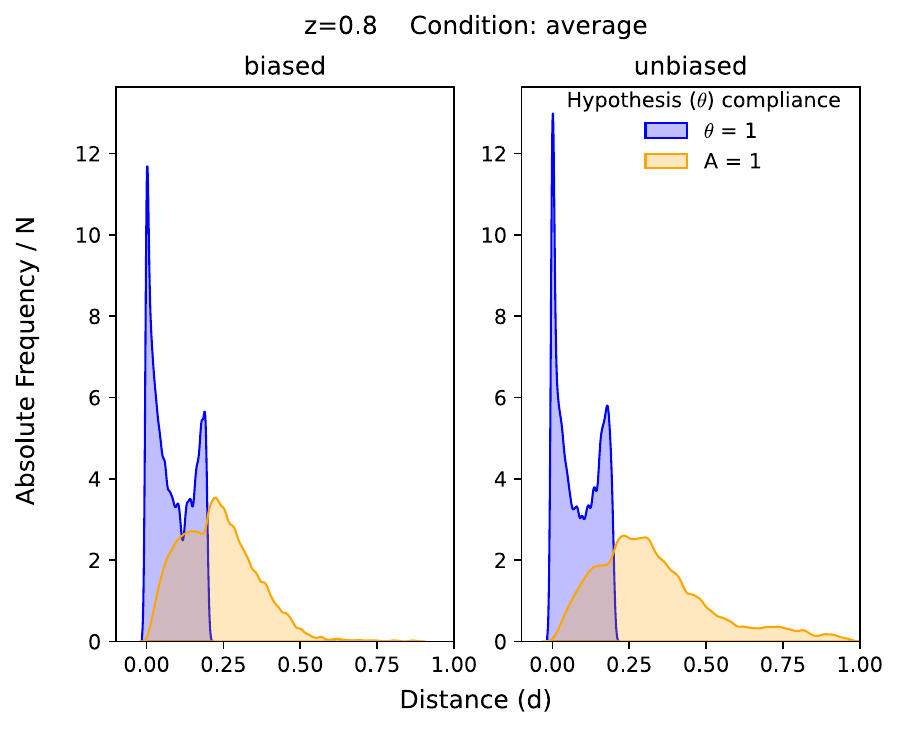}
    \caption{Distance density plots for $\theta_3$ at different thresholds for the biased and unbiased models. The models behave similarly with respect to this hypothesis.}
\label{fig:squad_h3_DD}
\end{figure}

To further corroborate our conclusions, we add a histogram to compare the amount of contribution of the first sentence of each model, across different data points (Figure \ref{fig:squad_contribution_hist}). As we expected, the contribution of the first sentence for the biased model is consistently more than 50\% of the total,
while for the unbiased model the first sentence's contribution is most often between 10 and 20\%. We also observe that sometimes the contribution of the first sentence is higher then 90\% for both models, and we expect those data points to be the ones which have the right answer in the first sentence.
 Finally, we add  plots to display the density of distances for $\theta_2$ (Figure \ref{fig:squad_h2_DD}) and $\theta_3$ (Figure \ref{fig:squad_h3_DD}) at different thresholds. As it emerges from the plots for $\theta_2$, the biased model has a lot of overlap between the compliant and non-compliant data points, while the unbiased model exhibits clearer differentiation between the  peaks as the threshold increases. This again confirms that the biased model is not able to discriminate nor to focus its contribution across different sentences, while the unbiased model can do both. As for the other plots, the two models manifest similar behaviours for $\theta_3$.
 

\end{document}